%% file: main.tex
\documentclass[10pt,twocolumn,letterpaper]{article}

\usepackage[pagenumbers]{cvpr} 

\input{preamble}

\definecolor{cvprblue}{rgb}{0.21,0.49,0.74}
\usepackage[pagebackref,breaklinks,colorlinks,citecolor=cvprblue]{hyperref}
\usepackage[normalem]{ulem}
\useunder{\uline}{\ul}{}
\usepackage{boldline}
\usepackage{graphicx}
\usepackage{multirow}
\usepackage{stfloats}

\title{GSFix3D: Diffusion-Guided Repair of Novel Views in Gaussian Splatting}

\author{
Jiaxin Wei$^{1}$\hspace{.4cm}
Stefan Leutenegger$^{1,2}$\hspace{.4cm}
Simon Schaefer$^{1}$
\\
$^{1}$Technical University of Munich\hspace{.3cm}
$^{2}$ETH Zurich
}

\begin{document}
\maketitle
\input{sec/0_abstract}    
\input{sec/1_intro}
\input{sec/2_related_work}
\input{sec/3_method}
\input{sec/4_exp}
\input{sec/5_conclusion}

\input{sec/6_acknowledgement}
{
    \small
    \bibliographystyle{ieeenat_fullname}
    \bibliography{main}
}
\input{sec/X_suppl}

\end{document}

%% file: preamble.tex
\usepackage[dvipsnames]{xcolor}


%% file: sec/0_abstract.tex
\begin{abstract}

Recent developments in 3D Gaussian Splatting have significantly enhanced novel view synthesis, yet generating high-quality renderings from extreme novel viewpoints or partially observed regions remains challenging. Meanwhile, diffusion models exhibit strong generative capabilities, but their reliance on text prompts and lack of awareness of specific scene information hinder accurate 3D reconstruction tasks. To address these limitations, we introduce \textbf{GSFix3D}, a novel framework that improves the visual fidelity in under-constrained regions by distilling prior knowledge from diffusion models into 3D representations, while preserving consistency with observed scene details. At its core is \textbf{GSFixer}, a latent diffusion model obtained via our customized fine-tuning protocol that can leverage both mesh and 3D Gaussians to adapt pretrained generative models to a variety of environments and artifact types from different reconstruction methods, enabling robust novel view repair for unseen camera poses. Moreover, we propose a random mask augmentation strategy that empowers GSFixer to plausibly inpaint missing regions. Experiments on challenging benchmarks demonstrate that our GSFix3D and GSFixer achieve state-of-the-art performance, requiring only minimal scene-specific fine-tuning on captured data. Real-world test further confirms its resilience to potential pose errors. Our code and data will be made publicly available. Project page: \href{https://gsfix3d.github.io}{gsfix3d.github.io}.

 
\end{abstract}

%% file: sec/1_intro.tex
\section{Introduction}

3D Gaussian Splatting (3DGS)~\cite{3dgs} has recently emerged as an efficient and expressive explicit representation that models scenes using a set of 3D Gaussian primitives and enables photorealistic rendering through differentiable rasterization. Compared to previous Neural Radiance Fields (NeRF)~\cite{nerf} approaches, it achieves faster convergence and significantly higher rendering speeds. However, a key limitation persists in those optimization-based representations as they heavily rely on meticulously curated and densely sampled input views to achieve high visual fidelity near the training camera poses. In regions with sparse observations or from viewpoints that deviate substantially from the training data, 3DGS struggles to infer plausible geometry and appearance, often producing artifacts such as incomplete surfaces, unnatural geometry, or visible holes that severely degrade image quality. Moreover, obtaining sufficient coverage and accurate measurements often requires labor-intensive data collection, costly high-end 3D scanners, and skilled operators, which largely limits the accessibility of such methods for casual users with only mobile devices.

In parallel, text-to-image generative models based on latent-space denoising diffusion, such as Stable Diffusion~\cite{stablediffusion}, have shown the remarkable ability to synthesize diverse and photorealistic images. Trained on large-scale, captioned images from the internet, those models effectively gain a deep understanding of 2D visual concepts. To obtain greater control over diffusion model outputs, a variety of techniques, such as ControlNet~\cite{controlnet}, T2I-Adapters~\cite{t2i}, and LoRA~\cite{lora}, have been proposed. Though powerful, these methods are primarily designed for image generation rather than repairing, and thus often lack input-output consistency, making them unsuitable for direct integration into 3D reconstruction pipelines where spatial and visual fidelity are critical.

To combine the strengths of diffusion models with existing 3D reconstructions, we introduce a novel view repair framework, GSFix3D, tailored for 3D Gaussian Splatting. Our method renders novel view images from initial reconstructions and refines them using a scene-adapted latent diffusion model by removing rendering artifacts and completing missing content. These enhanced images are then treated as pseudo-inputs and lifted back into 3D space to improve the underlying reconstruction. The key to our pipeline is a dedicated fine-tuning strategy that enables the pretrained diffusion model to internalize scene-specific priors, model artifact patterns, and develop inpainting capabilities using our proposed random mask augmentation. In contrast to DIFIX~\cite{difix}, which relies on large-scale curated real image pairs for training yet still lacks inpainting capabilities and struggles with unseen artifacts, our method requires only a one-time pretraining on two small synthetic datasets~\cite{hypersim, vkitti} to obtain a general base model. This base model can then be efficiently fine-tuned on the same captured data used for initial reconstruction, enabling adaptation to diverse scenes. The resulting module, GSFixer, acts as a plug-and-play image enhancer, transforming imperfect renderings into high-quality, photorealistic images.
Our main contributions are as follows:

\begin{itemize}
\item  We propose GSFix3D, a new pipeline for repairing novel views in 3DGS reconstructions that leverages the diffusion model, GSFixer, to enhance under-constrained regions. We exploit the complementary properties between 3DGS and traditional mesh representations to further boost repairing performance.

\item We introduce a customized fine-tuning protocol for pretrained diffusion models tailored to the novel view repair task. This protocol efficiently adapts the model to diverse scenes and reconstruction pipelines and enables it to internalize scene-specific priors, learn artifact patterns, and develop strong inpainting capabilities through our proposed random mask augmentation.

\item Experiments on challenging benchmarks demonstrate state-of-the-art performance under extreme novel viewpoints, with only a few hours of fine-tuning on the same captured data used for reconstruction using a single consumer GPU. Additional tests on self-collected real-world data further validate its robustness to pose inaccuracies. We will release the real-world data and selected extreme novel views from the Replica dataset~\cite{replica}.
\end{itemize}


%% file: sec/2_related_work.tex
\section{Related Work}

\subsection{3D Reconstruction and Mapping}

Traditional dense reconstruction methods, such as KinectFusion~\cite{kinectfusion}, fuse per-frame depth maps into a volumetric grid. Follow-up work improves scalability by using efficient data structures like octrees~\cite{steinbrucker2013large, supereight} and voxel hashing~\cite{voxelhasing, voxblox}. Though the reconstructed geometry suffices for robotics tasks such as navigation, it often lacks realism in visualization. Recently, NeRF~\cite{nerf} represents scenes as implicit neural functions. Several NeRF-based SLAM systems combine tracking and mapping within this framework~\cite{imap, niceslam}. Despite producing high-quality renderings, NeRF methods are computationally expensive and struggle with real-time applications. 3DGS~\cite{3dgs} addresses these limitations by representing scenes with explicit, differentiable Gaussian primitives, enabling faster rendering and optimization. This has led to several 3DGS-based SLAM systems: GS-SLAM~\cite{gsslam} uses opacity thresholds to drive adaptive Gaussian insertion, SplaTAM~\cite{splatam} employs a densification mask based on rendered silhouettes and depth, while MonoGS~\cite{monogs} relies on monocular depth estimates with variable uncertainty. To improve efficiency, RTG-SLAM~\cite{rtgslam} categorizes Gaussians as either opaque or transparent and updates only unstable ones, whereas GSFusion~\cite{gsfusion} integrates Truncated Signed Distance Field (TSDF)~\cite{tsdf} and 3DGS in a hybrid framework and employs a quadtree-based image segmentation strategy to reduce redundant splats. Despite these advances, challenges persist in handling under-constrained areas and achieving artifact-free reconstruction. We build our approach on 3DGS reconstructions due to their real-time performance, photorealistic rendering, and full differentiability, which make them particularly suitable for downstream repair tasks.

\subsection{Novel View Repair}

Although dense-view reconstruction has become increasingly reliable, novel view rendering remains susceptible to artifacts, especially in under-constrained regions. Prior work has largely focused on sparse-view settings, where such degradation is more obvious.~\cite{deceptivenerf} introduces a deceptive diffusion model that refines novel views rendered from few-view reconstructions and uses an uncertainty measure to improve consistency. RI3D~\cite{ri3d} uses two separate diffusion models for repairing visible regions and inpainting missing areas, whereas ours integrates these tasks into a single model. To improve temporal coherence, several methods leverage video diffusion models. 3DGS-Enhancer~\cite{3dgsenhancer} is the first to train a video diffusion model on a large-scale dataset created with pairs of low and high-quality images. GenFusion~\cite{genfusion} fine-tunes a video diffusion model on artifact-prone RGB-D videos using a masking strategy that simulates common view-dependent artifacts for content-aware outpainting, while~\cite{tamingvideo} uses training-free scene-grounding guidance to steer the video diffusion model toward temporally consistent synthesis. Despite promising results, these methods rely heavily on customized preprocessing steps to bootstrap initial reconstructions and carefully curated datasets to train diffusion models effectively. 

In this paper, we focus on novel view repair for reconstructions where artifacts still persist despite extensive coverage. SGD~\cite{sgd} introduces a tailored diffusion pipeline for autonomous driving scenarios, using adjacent frames as conditioning inputs and leveraging LiDAR point cloud to train a ControlNet for explicit depth control. DIFIX~\cite{difix} takes a step toward general view repair by training a single-step diffusion model on a large curated dataset of real noisy–clean image pairs, created via handcrafted corruption strategies. However, its performance drops when exposed to unseen artifacts and it struggles with inpainting. In contrast, our GSFixer is obtained through a lightweight fine-tuning protocol. With minimal pretraining on synthetic data and fine-tuning on captured reconstruction data, GSFixer achieves robust artifact removal, adapts to diverse pipelines and scenes, and exhibits strong inpainting capabilities, all within a single model that runs efficiently on consumer hardware.

%% file: sec/3_method.tex
\begin{figure*}[]
  \centering
   \includegraphics[width=\linewidth]{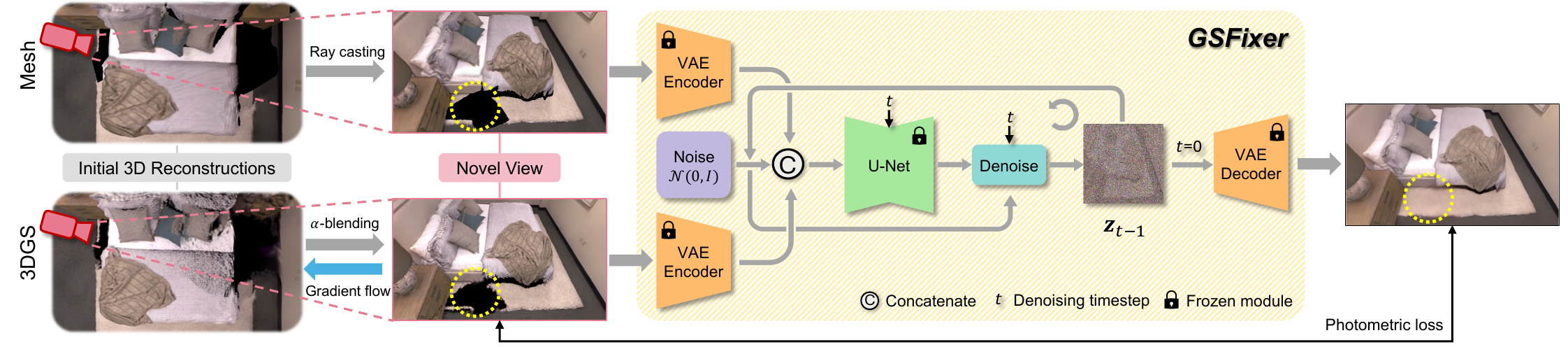}
   \caption{System overview of the proposed GSFix3D framework for novel view repair. Given initial 3D reconstructions in the form of mesh and 3DGS, we render novel views and use them as conditional inputs to GSFixer. Through a reverse diffusion process, GSFixer generates repaired images with artifacts removed and missing regions inpainted. These outputs are then distilled back into 3D by optimizing the 3DGS representation using photometric loss.}
   \label{fig:pipeline}
\end{figure*}

\section{Method}

Our goal is to enhance the photorealism of novel views in reconstructed 3DGS scenes, especially for viewpoints distant from the original camera trajectories and suffer from limited observations. We present a customized fine-tuning protocol to adapt a pretrained diffusion model for artifact removal and view inpainting (Sec.~\ref{Sec:protocol}). We then describe our inference scheme (Sec.~\ref{Sec:inference}), and how the fine-tuned model integrates into the full pipeline to improve the visual quality of novel views (Sec.~\ref{Sec:gsfix}). An overview of our method is illustrated in Fig.~\ref{fig:pipeline}.

\subsection{Fine-Tuning Protocol}
\label{Sec:protocol}

Given a reconstructed 3DGS scene, we formulate the image repair task as a conditional generation problem and fine-tune a pretrained latent diffusion model, i.e. Stable Diffusion v2~\cite{stablediffusion}, to learn the conditional distribution $p(I^{gt}|I^{c})$ where $I^{gt} \in \mathbb{R}^{H \times W \times 3}$ denotes the ground truth RGB image and $I^{c} \in \mathbb{R}^{H \times W \times 3}$ is the condition image rendered from the imperfect reconstruction.

\begin{figure}[]
  \centering
   \includegraphics[width=\linewidth]{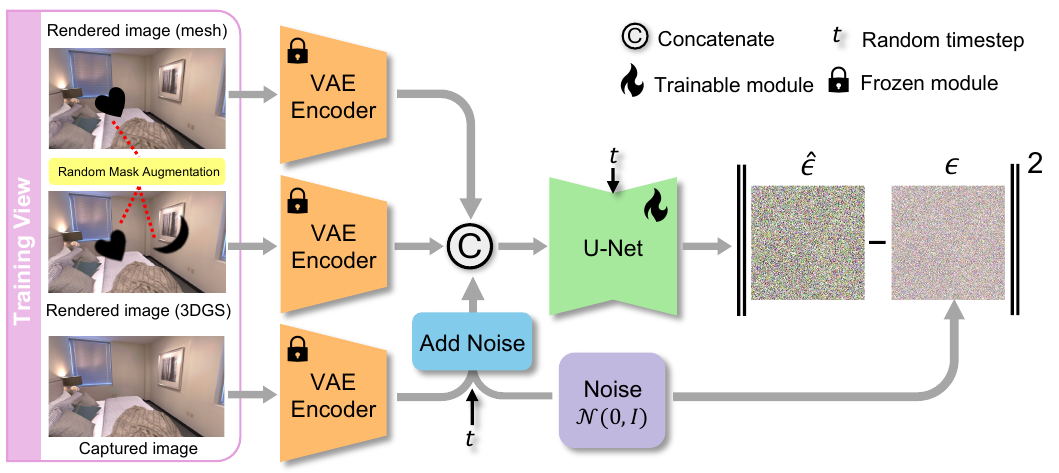}
   \caption{Illustration of the customized fine-tuning protocol for adapting a pretrained diffusion model into GSFixer, enabling it to handle diverse artifact types and missing regions.}
   \label{fig:finetuning_protocol}
\end{figure}

In our approach, we further extend the conditioning input to two rendered images: one from the 3D Gaussian Splatting representation ($I^{gs}$) and another from a mesh representation ($I^{mesh}$). Thus, the actual conditional distribution becomes $p(I^{gt}|I^{mesh}, I^{gs})$. This dual-conditioning strategy is motivated by the complementary strengths of 3DGS and traditional mesh-based reconstructions. 3DGS, as an optimization-based method, tends to suffer in regions with sparse observations, often leading to visible artifacts such as holes or incomplete geometry. Mesh reconstructions, though usually less photorealistic at lower resolutions, offer more coherent geometry and stronger spatial priors in under-constrained areas. By jointly leveraging both representations, we aim to provide the diffusion model with richer appearance cues for image refinement. To ensure that the mesh input remains geometrically consistent yet independent from the 3DGS optimization process, we obtain the mesh and the 3DGS map simultaneously using GSFusion~\cite{gsfusion}, an online RGB-D mapping system. This avoids directly extracting the mesh from the 3DGS representation, as done in prior works~\cite{sugar, 2dgs}, which could introduce correlated artifacts. The overall fine-tuning protocol is presented in \cref{fig:finetuning_protocol}. We conduct an ablation study in \cref{Sec:ablation} comparing the performance of using both inputs versus 3DGS alone, validating the effectiveness of our design choice.

\subsubsection{Network Architecture}
\label{Sec:network}

Diffusion models~\cite{dm2019, dm2021, stablediffusion} are a class of generative frameworks that generate data by learning to invert a progressively noised process. We use a frozen Variational Autoencoder (VAE)~\cite{vae} to encode all images into a latent space, enabling diffusion-based learning in a more compact domain. For a given image $I$, its latent code is obtained via the encoder $\mathcal{E}: \mathbf{z}=\mathcal{E}(I)$. This results in a latent triplet $(\mathbf{z}^{mesh}, \mathbf{z}^{gs}, \mathbf{z}^{gt})$. To train the denoising model, we follow the standard Denoising Diffusion Probabilistic Models (DDPM)~\cite{ddpm} formulation and incrementally add standard Gaussian noise $\epsilon \sim \mathcal{N}(0, \mathbf{I})$ to the clean ground-truth latent $\mathbf{z}_0 := \mathbf{z}^{gt}$ over $T$ discrete timesteps, producing a sequence $\{\mathbf{z}_t\}_{t=1}^{T}$. The noisy latent at timestep $t$ is then given by:
\begin{equation}
    \mathbf{z}_t = \sqrt{\bar{\alpha}_t} \mathbf{z}_0 + \sqrt{1 - \bar{\alpha}_t} \epsilon, \label{Eq:noised_latent}
\end{equation}
where $\bar{\alpha}_t$ denotes the cumulative product of noise schedule coefficients~\cite{ddpm, ddim}. Following~\cite{marigold}, we repurpose the U-Net backbone from the pretrained diffusion model into a conditional denoiser for image repair. We concatenate the latent codes along the feature dimension to form the input $\bar{\mathbf{z}}_t=\text{concat}(\mathbf{z}^{mesh}, \mathbf{z}^{gs}, \mathbf{z}_t)$. To accommodate the increased channel count, we expand the first layer of the U-Net by duplicating the original weight tensor and dividing its values by three. This design choice maintains the original weight distribution and prevents excessive activation scaling, allowing us to preserve the initialization behavior of the pretrained model while enabling conditional inputs. The conditional U-Net $\epsilon_\theta$ is then trained to predict the added noise by minimizing a standard DDPM objective:
\begin{equation}
\mathcal{L}= \mathbb{E}_{\mathbf{z}_0, \epsilon \sim \mathcal{N}(0, \mathbf{I}), t \sim \mathcal{U}[1, T]} \left[ \left\| \epsilon -  \hat{\epsilon} \right\|^2 \right],
\end{equation}
where $\hat{\epsilon}=\epsilon_\theta(\bar{\mathbf{z}}_t, t)$ is the predicted noise.

\subsubsection{Data Augmentation}
To construct our training set, we render each captured view using both the mesh and the 3DGS map, resulting in paired triplets $(I^{mesh}, I^{gs}, I^{gt})$, where $I^{gt}$ is the original captured RGB image, $I^{gs}$ is the image rendered from 3DGS via $\alpha$-blending, and $I^{mesh}$ is obtained via ray-casting on the mesh. This process requires no additional data beyond the original captured RGB images, their corresponding camera poses, and the reconstructed maps.

Direct fine-tuning on these triplets can already help the diffusion model adapt to the scene and learn to remove specific artifacts in the 3DGS rendering. However, one major challenge remains: the model's ability to inpaint missing regions, which usually appear as black holes in novel views due to under-constrained geometry or occlusions. Since all training images are rendered from the original captured viewpoints, they are mostly complete in appearance and fail to expose the model to such corner cases.

To explicitly train the model to handle incomplete renderings, we introduce a masking-based data augmentation scheme. For each training triplet, we randomly select a semantic mask from a set of annotated real-image masks~\cite{pipe_masks}. The key intuition is to leverage the diverse mask shapes derived from real-world object semantics, which not only enhances the realism of the masked regions but also eliminates the need for manually designing complex rules to simulate missing areas caused by various factors such as occlusion or under-constrained observations. This mask is applied in two distinct ways: (1) the same mask is overlaid on both $I^{mesh}$ and $I^{gs}$, simulating occlusions that might occur in novel views; and (2) an additional, independent mask is applied solely to $I^{gs}$ to simulate the common degradation of 3DGS renderings in regions with limited observations. To better approximate the soft boundaries in 3DGS renderings, we further apply a small amount of Gaussian blur to the mask used on $I^{gs}$. We evaluate the impact of this augmentation strategy in \cref{Sec:ablation}, where we compare models trained with and without random masks and show its importance in improving the inpainting ability for novel views.

\subsection{Inference with GSFixer}
\label{Sec:inference}

At inference time, we freeze the fine-tuned U-Net parameters and apply the model to novel views, as illustrated in \cref{fig:pipeline}. We begin by encoding the conditional inputs, i.e., the rendered images from novel viewpoints, into the latent space using the frozen VAE encoder. The latent for the target image to be generated, $\mathbf{z}_t$, is initialized as standard Gaussian noise. We then concatenate these latent codes in the same order used during fine-tuning to form the diffusion model input: $\bar{\mathbf{z}}_t=\text{concat}(\mathbf{z}^{mesh}, \mathbf{z}^{gs}, \mathbf{z}_t)$. To generate the fixed image, we iteratively denoise $\mathbf{z}_t$ using the deterministic Denoising Diffusion Implicit Model (DDIM)~\cite{ddim} schedule to perform efficient non-Markovian sampling. The update at each timestep is as follows:
\begin{equation}
\mathbf{z}_{t-1} = \sqrt{\bar{\alpha}_{t-1}} \hat{\mathbf{z}}_0 + \sqrt{1 - \bar{\alpha}_{t-1}} \epsilon_\theta(\bar{\mathbf{z}}_t, t),
\end{equation}
where the clean latent $\hat{\mathbf{z}}_0$ is estimated as:
\begin{equation}
\hat{\mathbf{z}}_0 = \frac{1}{\sqrt{\bar{\alpha}_t}} \left( \mathbf{z}_t - \sqrt{1 - \bar{\alpha}_t} \epsilon_\theta(\bar{\mathbf{z}}_t, t) \right),
\end{equation}
derived directly from the forward diffusion formulation in \cref{Eq:noised_latent}. After completing the denoising process, the final fixed image is obtained by decoding the predicted clean latent using the VAE decoder $\mathcal{D}: \hat{I}^{fixed}=\mathcal{D}(\mathbf{z}_0)$.

\subsection{GSFix3D: Diffusion-Guided Novel View Repair}
\label{Sec:gsfix}

The final stage of our GSFix3D framework lifts the output of the diffusion model, i.e., GSFixer, back into the 3D representation. Thanks to the full differentiability of 3DGS, we can continue optimizing the parameters of the initial 3DGS reconstruction by minimizing a photometric loss between the fixed image $\hat{I}^{fixed}$ and the rendered image $I^{gs}$:
\begin{equation*}
    \mathcal{L}_{\text{pho}} = (1 - \lambda) \|\hat{I}^{fixed} - I^{gs}\|_1 + \lambda \mathcal{L}_{\text{SSIM}}(\hat{I}^{fixed}, I^{gs}),
\end{equation*}
where $\lambda$ is a weighting factor, and $\mathcal{L}_{\text{SSIM}}$ denotes the Structural Similarity loss. We also enable adaptive density control during optimization, following~\cite{3dgs}, to fill in previously empty or under-populated regions.

To reduce inconsistencies in the repaired images and improve global coherence, we further append the repaired views and their corresponding poses to the original captured datasets and then optimize over this augmented dataset for several iterations. Note that we use a sparse set of keyframes recorded during the initial reconstruction phase instead of the full dataset to avoid redundant and time-consuming optimization.

%% file: sec/4_exp.tex
\section{Experiments}

\subsection{Experimental Setup}

\noindent \textbf{Evaluation Datasets and Metrics.} We compare different methods on two challenging benchmark datasets: ScanNet++\cite{scannetpp} and Replica\cite{replica}. ScanNet++ is a real-world indoor dataset containing high-quality RGB-D data. Each scene includes two separate camera trajectories for training and evaluation, respectively. Following \cite{gsfusion}, we select four scenes from ScanNet++: 8b5caf3398, 39f36da05b, b20a261fdf, and f34d532901. The Replica dataset consists of photorealistic synthetic indoor scenes with accurate RGB-D imagery. We use consistent trajectories from \cite{niceslam} for reconstruction and fine-tuning. To enable the quantitative assessment of novel views and to evaluate inpainting capabilities, we manually render ground truth novel views from extreme viewpoints with large unobserved regions (see \cref{Sec:replica_data} in the supplementary). We use three common metrics to measure rendering quality and fidelity: PSNR, SSIM, and LPIPS. All reported results are averaged over scenes within each dataset.

\noindent \textbf{Baselines.} We compare our GSFixer against two variants from~\cite{difix}: DIFIX and DIFIX-ref. DIFIX is a single-step image diffusion model trained on 80k noisy-clean image pairs curated from real-world datasets. DIFIX-ref extends this setup by incorporating an additional reference view as input, introducing multi-view constraints to enhance performance. In addition to GSFusion, we also include two recent Gaussian SLAM methods, SplaTAM\cite{splatam} and RTG-SLAM\cite{rtgslam}, as alternative sources of 3D reconstructions, each exhibiting distinct artifact patterns due to differences in initialization and optimization strategies. We apply the above image repair models to novel view renderings produced by each of these reconstruction methods.

\begin{table*}[]
\centering
\resizebox{0.62\linewidth}{!}{%
\begin{tabular}{l|ccc|ccc}
\hlineB{2}
\multirow{2}{*}{Method}      & \multicolumn{3}{c|}{ScanNet++}                      & \multicolumn{3}{c}{Replica}                         \\ \cline{2-7} 
                             & PSNR$\uparrow$ & SSIM$\uparrow$ & LPIPS$\downarrow$ & PSNR$\uparrow$ & SSIM$\uparrow$ & LPIPS$\downarrow$ \\ \hlineB{2}
SplaTAM                      & 23.03          & 0.791          & 0.311             & {\ul 23.82}    & {\ul 0.833}    & 0.267             \\
SplaTAM + DIFIX              & {\ul 23.06}    & 0.789          & 0.220             & 22.97          & 0.790          & 0.262             \\
SplaTAM + DIFIX-ref          & 22.79          & {\ul 0.799}    & {\ul 0.203}       & 22.97          & 0.830          & {\ul 0.217}       \\
SplaTAM + GSFixer            & \textbf{25.11} & \textbf{0.831} & \textbf{0.188}    & \textbf{25.67} & \textbf{0.839} & \textbf{0.215}    \\ \hline
RTG-SLAM                     & {\ul 19.54}    & {\ul 0.777}    & 0.341             & {\ul 25.00}    & \textbf{0.860} & 0.247             \\
RTG-SLAM + DIFIX             & 19.43          & 0.762          & 0.245             & 24.02          & 0.811          & {\ul 0.214}       \\
RTG-SLAM + DIFIX-ref         & 19.29          & 0.769          & {\ul 0.223}       & 23.89          & 0.834          & \textbf{0.193}    \\
RTG-SLAM + GSFixer           & \textbf{24.80} & \textbf{0.824} & \textbf{0.204}    & \textbf{26.27} & {\ul 0.843}    & 0.228             \\ \hline
GSFusion (gs)                & 24.58          & \textbf{0.838} & 0.308             & 22.10          & {\ul 0.844}    & 0.296             \\
GSFusion (gs) + DIFIX        & 24.34          & 0.818          & 0.193             & 21.81          & 0.772          & 0.273             \\
GSFusion (gs) + DIFIX-ref    & 23.83          & 0.822          & {\ul 0.184}       & 21.91          & 0.821          & {\ul 0.224}       \\
GSFusion (gs) + GSFixer      & {\ul 24.79}    & 0.833          & 0.196             & {\ul 23.87}    & 0.830          & 0.251             \\
GSFusion (mesh+gs) + GSFixer & \textbf{25.30} & {\ul 0.837}    & \textbf{0.183}    & \textbf{25.98} & \textbf{0.845} & \textbf{0.219}    \\ \hlineB{2}
\end{tabular}%
}
\caption{Comparisons of diffusion-based repair methods on the ScanNet++ and Replica datasets. The best result is highlighted in \textbf{bold}, and the second-best is \uline{underlined}. The text inside ( ) indicates the format of the reconstruction used.}
\label{tab:main_exp}
\end{table*}

\noindent \textbf{Implementation Details.} We adopt Stable Diffusion v2 as our base latent diffusion model, disabling text prompt and applying the fine-tuning protocol described in \cref{Sec:protocol}. During training, we use the DDPM noise scheduler with 1000 diffusion steps. For inference, we follow the DDIM scheduler with only 4 steps for accelerated sampling. Considering the difficulty of collecting large-scale real-world training pairs for this task, we first pretrain the modified U-Net (see \cref{Sec:network}) for 6k iterations on two synthetic datasets: Hypersim\cite{hypersim} for indoor scenes and Virtual KITTI~\cite{vkitti} for outdoor street environments. We use a batch size of 2 and accumulate gradients over 16 steps to stabilize training with the Adam optimizer. The learning rate is set to $3\times 10^{-5}$. We acquire the geometrically aligned mesh and 3DGS map by running GSFusion and fine-tune the pretrained model separately for each scene. For real scenes from ScanNet++, we fine-tune for 800 iterations. For synthetic scenes from Replica, we fine-tune for 400 iterations. The fine-tuning process typically takes 4 hours for ScanNet++ and 2 hours for Replica. As for 3DGS optimization in GSFix3D, we perform 20 iterations for each repaired image and 50 iterations over the augmented dataset. All experiments are conducted on a single NVIDIA RTX 4500 Ada GPU with 24GB VRAM.

\begin{figure*}[]
  \centering
   \includegraphics[width=\linewidth]{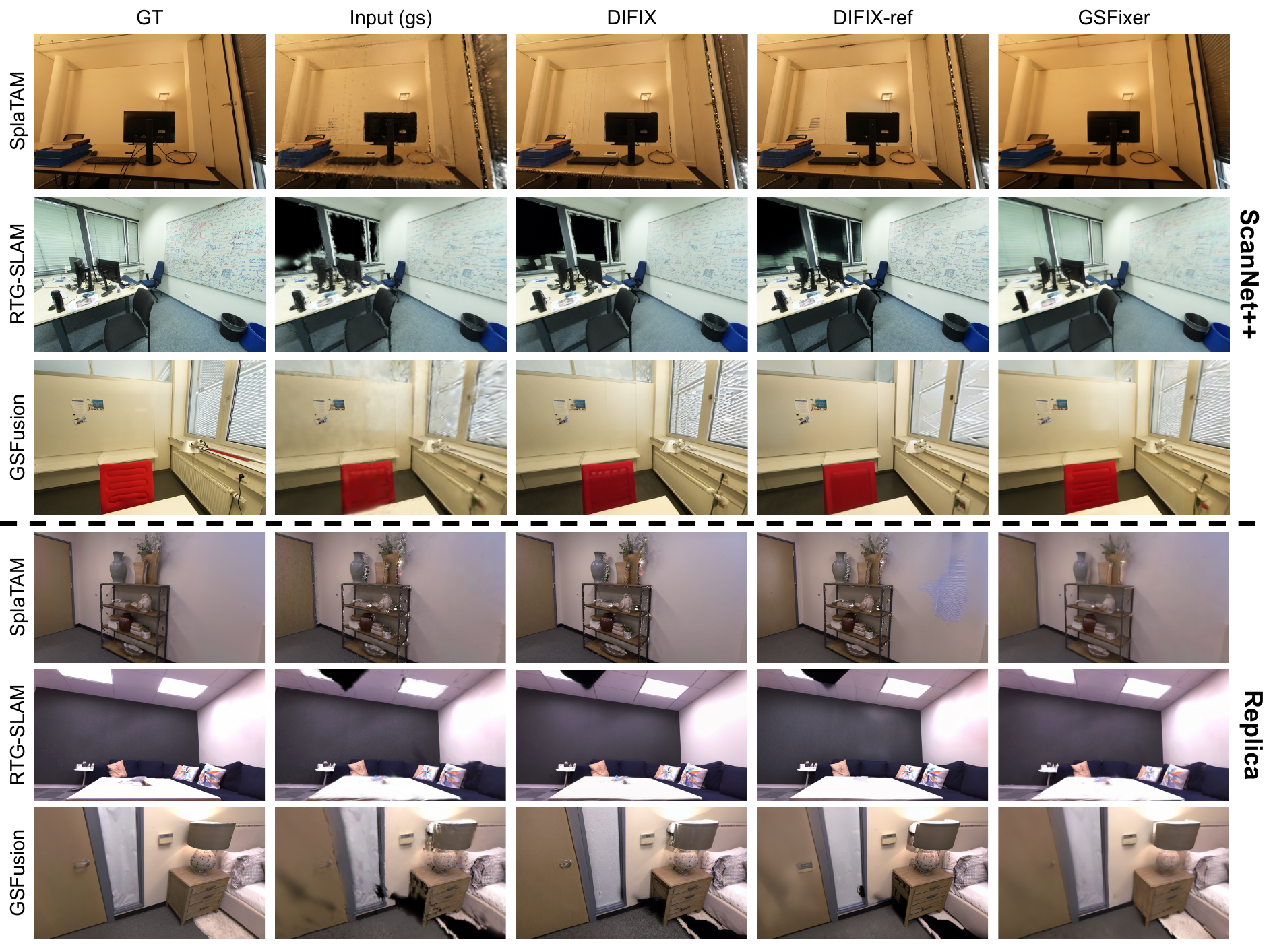}
   \caption{Qualitative comparisons of diffusion-based repair methods on the ScanNet++ and Replica datasets. All examples use only 3DGS reconstructions as the input source. Our GSFixer effectively removes artifacts and fills in large holes, where both DIFIX and DIFIX-ref fail to produce satisfactory results.}
   \label{fig:main_exp_vis}
\end{figure*}

\begin{figure*}[]
  \centering
   \includegraphics[width=\linewidth]{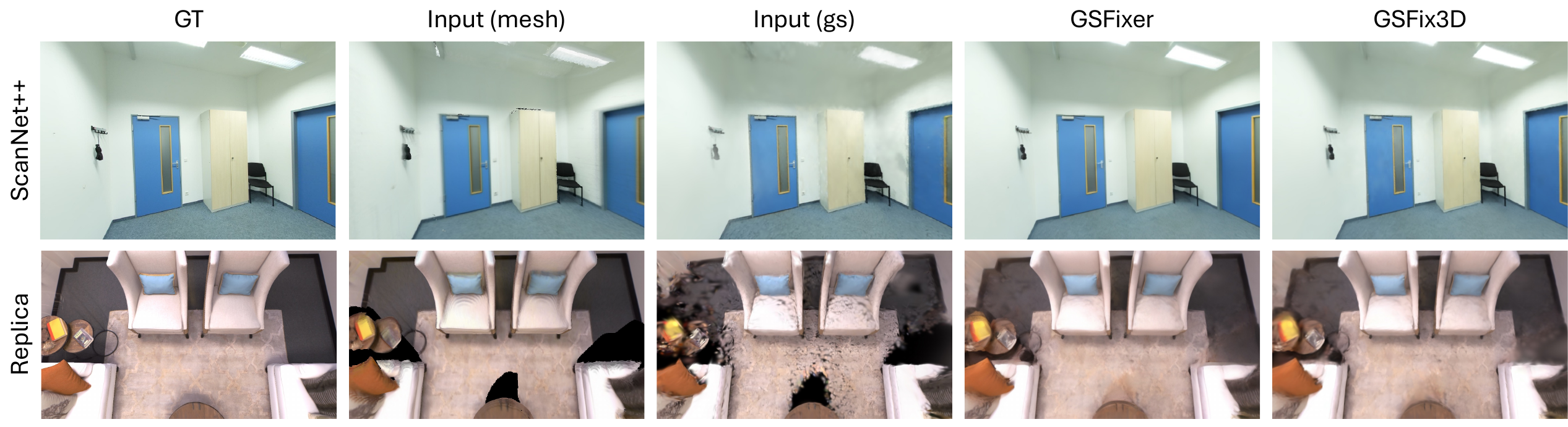}
   \caption{Qualitative comparison between GSFixer and GSFix3D on the ScanNet++ and Replica datasets. Both mesh and 3DGS reconstructions from GSFusion are used as input sources. The 2D visual improvements from GSFixer are effectively distilled into the 3D space by GSFix3D.}
   \label{fig:main_exp_gsfix_vis}
\end{figure*}

\begin{figure*}[]
  \centering
   \includegraphics[width=0.9\linewidth]{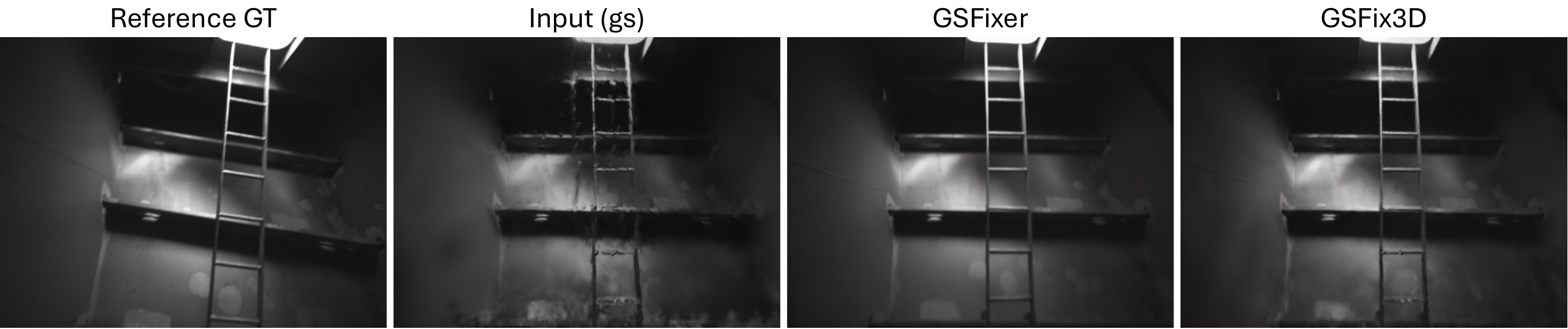}
   \caption{Novel view repair on self-collected ship data. Our method is robust to pose errors, effectively removing shadow-like floaters.}
   \label{fig:ship_cam0_vis}
\end{figure*}

\subsection{Results}
\label{Sec:results}

\Cref{tab:main_exp} reports quantitative results on ScanNet++ and Replica. For SplaTAM and RTG-SLAM, which output only 3DGS maps, we fine-tune GSFixer exclusively on rendered images from their reconstructions. Despite this constraint, GSFixer consistently outperforms DIFIX and DIFIX-ref across all metrics on ScanNet++, with over 5 dB PSNR gain in the \textit{RTG-SLAM+GSFixer} setting. Qualitative results in \cref{fig:main_exp_vis} show that, in the RTG-SLAM example, DIFIX and DIFIX-ref leave a large black hole where a window is missing, while GSFixer fills it with plausible content. In the SplaTAM example, baselines leave colorful floaters, whereas GSFixer learns their patterns and removes them.


For GSFusion, which provides both a mesh and a 3DGS map, we introduce a dual-input setting, \textit{GSFusion(mesh+gs)+GSFixer}, that further boosts performance over the single-input variant. We analyze this effect in detail in \cref{Sec:ablation}. On the more challenging Replica dataset, where we evaluate on manually selected extreme novel viewpoints with large unobserved regions, GSFixer again outperforms baselines in PSNR and remains competitive in SSIM. The strong inpainting ability of GSFixer is visually evident on the Replica dataset in \cref{fig:main_exp_vis}. Interestingly, DIFIX and DIFIX-ref achieve lower LPIPS scores in some cases, which we attribute to their sharp visual details. This is likely due to their training on 80k noisy-clean image pairs curated from real-world datasets (though the dataset is not publicly available), whereas GSFixer is only pretrained on two synthetic datasets and fine-tuned on a limited amount of clean captured data. We explore additional comparisons and results in \cref{Sec:more_difix_results} in the supplementary.


The overall performance of our GSFix3D framework is reported in \cref{tab:gsfix_exp}. Compared to the direct outputs from GSFixer, lifting the repaired images back into the 3D representation leads to improved perceptual quality thanks to multi-view constraints, as evidenced by higher PSNR and SSIM scores. However, due to the optimization characteristics of the 3DGS representation, the final renderings tend to be less smooth than the 2D generative results, which accounts for the slightly higher LPIPS values. Qualitative examples are provided in \cref{fig:main_exp_gsfix_vis}. We further apply the full GSFix3D framework to SplaTAM and RTG-SLAM reconstructions, demonstrating its effectiveness in \cref{Sec:more_gsfix_results} of the supplementary material.


\subsection{Real-World Evaluation in the Wild}

We collect a stereo sequence inside a ship structure using an Intel RealSense D455 camera. We compute depth maps for the left camera using FoundationStereo~\cite{foundationstereo} for improved quality, and estimate camera poses with OKVIS2~\cite{okvis2}. Since no ground truth is available, the estimated poses may contain errors. Those post-processed data are then fed into GSFusion to obtain an initial 3DGS reconstruction. We fine-tune a GSFixer model using 3DGS renderings as input. \Cref{fig:ship_cam0_vis} shows a novel view example where shadow-like floaters appear near the ladder due to inaccurate poses. Our method effectively removes these artifacts in 2D and distills the correction back into the 3D representation, demonstrating robustness to common pose errors in real-world data collection, particularly in uncontrolled settings without high-end equipment or precise calibration. Additional real-world results are provided in \cref{Sec:more_real_world} of the supplementary, including a test on an outdoor scene~\cite{fast-livo} using a LiDAR-Inertial-Camera Gaussian Splatting SLAM system~\cite{gaussian-lic}, which further demonstrates the practical adaptability of our method.

\begin{table}[]
\centering
\resizebox{\columnwidth}{!}{%
\begin{tabular}{l|l|ccc}
\hlineB{2}
Dataset                    & Method                       & PSNR$\uparrow$ & SSIM$\uparrow$ & LPIPS$\downarrow$ \\ \hlineB{2}
\multirow{3}{*}{ScanNet++} & GSFusion (gs)                & 24.58          & {\ul 0.838}    & 0.308             \\
                           & GSFusion (mesh+gs) + GSFixer & {\ul 25.30}    & 0.837          & \textbf{0.183}    \\
                           & GSFusion (mesh+gs) + GSFix3D & \textbf{25.63} & \textbf{0.845} & {\ul 0.238}       \\ \hline\hline
\multirow{3}{*}{Replica}   & GSFusion (gs)                & 22.10          & 0.844          & 0.296             \\
                           & GSFusion (mesh+gs) + GSFixer & {\ul 25.98}    & {\ul 0.845}    & \textbf{0.219}    \\
                           & GSFusion (mesh+gs) + GSFix3D & \textbf{26.49} & \textbf{0.864} & {\ul 0.252}       \\ \hlineB{2}
\end{tabular}%
}
\caption{Comparisons of GSFixer and GSFix3D on the ScanNet++ and Replica datasets.}
\label{tab:gsfix_exp}
\end{table}

\subsection{Ablation Studies}
\label{Sec:ablation}

\begin{figure*}[]
  \centering
   \includegraphics[width=0.9\linewidth]{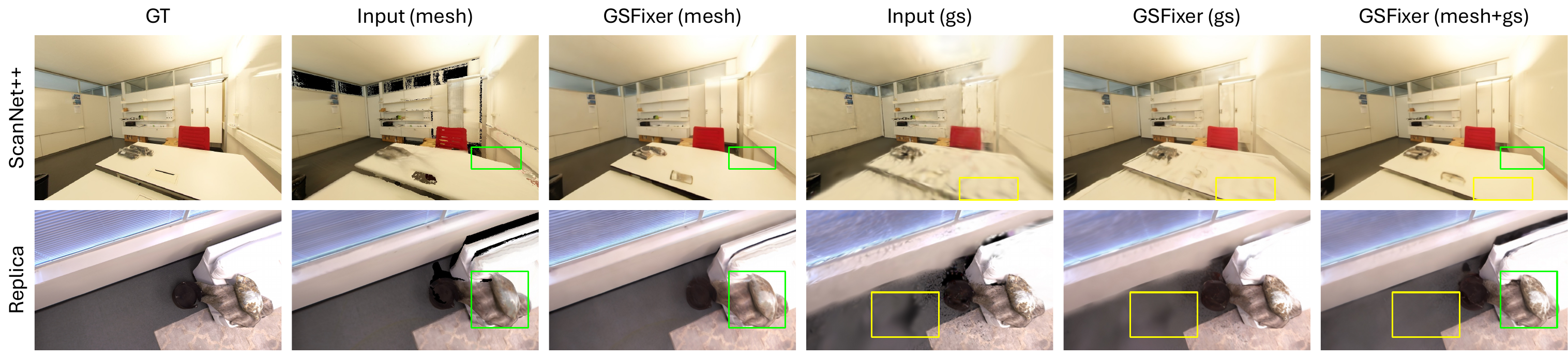}
   \caption{Qualitative ablation of input image conditions on the ScanNet++ and Replica datasets. We compare GSFixer results using three types of inputs rendered from GSFusion: mesh-only, 3DGS-only, and dual-input. The artifacts (highlighted by green and yellow boxes) present in the single-input settings are effectively mitigated with the dual-input configuration.}
   \label{fig:ablation_condition_vis}
\end{figure*}

\begin{figure*}[]
  \centering
   \includegraphics[width=0.9\linewidth]{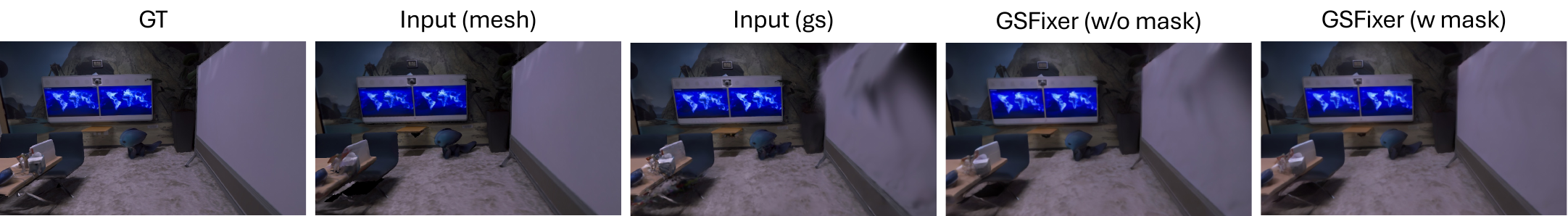}
   \caption{Qualitative ablation of random mask augmentation on the Replica dataset. We compare GSFixer results fine-tuned with and without our proposed augmentation strategy. The differences in inpainting quality highlight the improved ability to fill large missing regions when augmentation is used.}
   \label{fig:ablation_mask_vis}
\end{figure*}

\noindent \textbf{Image Conditions.} 
To analyze the impact of different input image conditions, we evaluate GSFixer under three input configurations: mesh-only, 3DGS-only, and dual-input (mesh+3DGS), with results in \cref{tab:ablation_condition}. On the synthetic Replica, which provides highly accurate measurements, mesh-based renderings tend to be of higher quality than their 3DGS counterparts from novel viewpoints. As a result, the \textit{GSFusion(mesh)+GSFixer} setting achieves better rendering performance than \textit{GSFusion(gs)+GSFixer}. In contrast, on the real-world ScanNet++ dataset, 3DGS reconstructions outperform mesh renderings due to noisy depth, making \textit{GSFusion(gs)+GSFixer} the better choice. When both images are used together as input, we observe complementary advantages: the dual-input setup leads to improved performance on ScanNet++ and a modest gain on Replica.

Qualitative results in \cref{fig:ablation_condition_vis} further highlight this benefit. For example, in a ScanNet++ scene, the mesh-rendered image suffers from geometric inaccuracies along the table edge, while the 3DGS-rendered image shows visual gaps on the table surface. When both are used to condition GSFixer, these issues are effectively mitigated. Similarly, in a Replica scene, the mesh-rendered image exhibits blurry textures on the pillow, and the 3DGS-rendered image contains visible holes on the floor. Combining both inputs allows GSFixer to resolve these artifacts by leveraging strengths from each source. Additional experiments on SplaTAM and RTG-SLAM are presented in \cref{Sec:more_ablation} of the supplementary.

\noindent \textbf{Random Mask Augmentation.} 
To validate the effectiveness of our proposed data augmentation strategy in improving inpainting capability for novel view repair, we conduct an ablation study by disabling the random mask augmentation during fine-tuning on the Replica dataset. We choose Replica for this evaluation due to its challenging novel views with extensive unobserved regions and visible holes. As shown in \cref{tab:ablation_augmentation}, GSFixer fine-tuned with random mask augmentation consistently outperforms the variant without augmentation across all metrics. It is also evident in \cref{fig:ablation_mask_vis}. The 3DGS-rendered image contains a large missing region on the whiteboard. Without random mask augmentation, GSFixer struggles to inpaint the hole even when given an additional mesh-rendered image as a condition. In contrast, our full model with augmentation successfully fills in the missing region with coherent and realistic textures, demonstrating its generalization to real occlusions.

\begin{table}[]
\centering
\resizebox{\columnwidth}{!}{%
\begin{tabular}{l|l|ccc}
\hlineB{2}
Dataset                    & Method                     & PSNR$\uparrow$ & SSIM$\uparrow$ & LPIPS$\downarrow$ \\ \hlineB{2}
\multirow{5}{*}{ScanNet++} & GSFusion (mesh)            & 17.87          & 0.750          & 0.358             \\
                           & GSFusion (mesh) + GSFixer    & 24.64          & 0.823          & 0.198             \\ \cline{2-5} 
                           & GSFusion (gs)              & 24.58          & \textbf{0.838} & 0.308             \\
                           & GSFusion (gs) + GSFixer      & {\ul 24.79}    & 0.833          & {\ul 0.196}       \\ \cline{2-5} 
                           & GSFusion (mesh+gs) + GSFixer & \textbf{25.30} & {\ul 0.837}    & \textbf{0.183}    \\ \hline \hline
\multirow{5}{*}{Replica}   & GSFusion (mesh)            & 23.20          & \textbf{0.849} & {\ul 0.217}       \\
                           & GSFusion (mesh) + GSFixer    & \textbf{26.61} & {\ul 0.846}    & \textbf{0.200}    \\ \cline{2-5} 
                           & GSFusion (gs)              & 22.10          & 0.844          & 0.296             \\
                           & GSFusion (gs) + GSFixer      & 23.87          & 0.830          & 0.251             \\ \cline{2-5} 
                           & GSFusion (mesh+gs) + GSFixer & {\ul 25.98}    & 0.845          & 0.219             \\ \hlineB{2}
\end{tabular}%
}
\caption{Ablation of image conditions on the ScanNet++ and Replica datasets.}
\label{tab:ablation_condition}
\end{table}

\begin{table}[]
\centering
\resizebox{0.8\columnwidth}{!}{%
\begin{tabular}{l|ccc}
\hlineB{2}
\begin{tabular}[c]{@{}l@{}}GSFusion (mesh+gs)\end{tabular} & PSNR$\uparrow$ & SSIM$\uparrow$ & LPIPS$\downarrow$ \\ \hlineB{2}
+ GSFixer (w/o mask)                                                  & 23.54          & 0.830          & 0.231             \\
+ GSFixer (w mask)                                                    & \textbf{25.98} & \textbf{0.845} & \textbf{0.219}    \\ \hlineB{2}
\end{tabular}
}
\caption{Ablation of random mask augmentation on the Replica dataset.}
\label{tab:ablation_augmentation}
\end{table}

%% file: sec/5_conclusion.tex
\section{Conclusion}


GSFix3D raises the bar for novel view repair in 3DGS reconstructions, requiring no massive real data curation or costly pertaining, only minimal fine-tuning on a small set of captured views. By coupling this efficient fine-tuning protocol with a dual-input design that fuses mesh and 3DGS cues, and empowering it with random mask augmentation as the key to strong inpainting performance, the resulting diffusion model, GSFixer, removes artifacts, fills missing regions with plausible detail, and adapts seamlessly to different scenes and reconstruction pipelines. Across diverse and challenging benchmarks, our method consistently outperforms prior diffusion-based approaches, validating its effectiveness, adaptability, and robustness even under pose inaccuracies, underscoring its practicality for a wide range of 3D reconstruction scenarios.

%% file: sec/6_acknowledgement.tex
\paragraph{Acknowledgement.} The authors gratefully acknowledge support from the EU project AUTOASSESS (Grant 101120732). We also thank Jaehyung Jung and Sebastián Barbas Laina for their assistance with ship data collection and processing, and Helen Oleynikova for her valuable feedback on the manuscript.

%% file: sec/X_suppl.tex
\clearpage
\setcounter{page}{1}
\maketitlesupplementary

\section{Data Preparation}

\subsection{Novel View Selection for Replica}
\label{Sec:replica_data}

\begin{figure*}[]
  \centering
   \includegraphics[width=\linewidth]{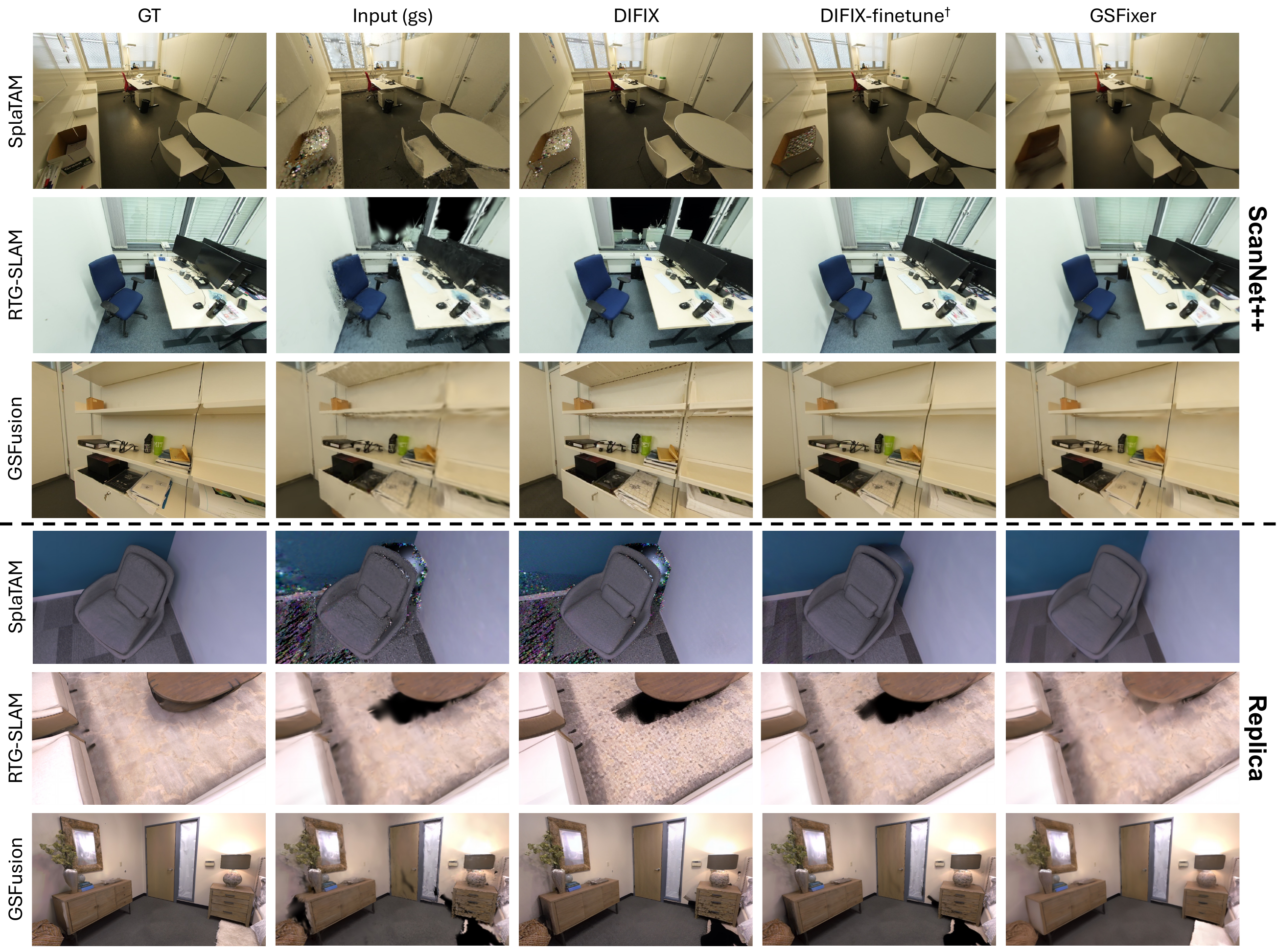}
   \caption{More qualitative comparisons of diffusion-based repair methods on the ScanNet++ and Replica datasets. All examples use only 3DGS reconstructions as the input source. Zoom in to better observe how GSFixer effectively removes artifacts and fills in large holes, where both DIFIX and DIFIX-finetune$^\dagger$ fail to produce satisfactory results.}
   \label{fig:main_exp_vis_supp}
\end{figure*}

The Replica dataset~\cite{replica} contains high-quality reconstructions of diverse indoor scenes, featuring clean dense geometry and high-resolution textures. We leverage the provided 3D models and the official Replica SDK to render novel view images, which serve as ground truth for quantitative evaluation in \cref{Sec:results}. We adopt the same camera intrinsics and image resolution ($1200\times 680$) as the training views provided by~\cite{niceslam} when generating novel views. Since these trajectories do not cover the entire scene, certain areas in the reconstructed map remain unobserved or under-constrained, often exhibiting artifacts. To assess inpainting performance, we deliberately select novel views that include such missing or artifact-prone regions, where existing pipelines (e.g., SplaTAM, RTG-SLAM, GSFusion) typically fail. This results in an intentionally challenging novel view repair task. Examples of selected novel views from the Replica dataset are shown in \cref{fig:main_exp_vis}, \cref{fig:main_exp_gsfix_vis}, \cref{fig:ablation_condition_vis}, \cref{fig:ablation_mask_vis}, \cref{fig:main_exp_vis_supp}, \cref{fig:ablation_condition_vis_supp} and \cref{fig:gsfix_vis_supp}. We will release the curated Replica novel view dataset to support reproducible research.

\subsection{Real-World Data Collection}
\label{Sec:real_data}

We collect a stereo image sequence inside a ship’s ballast water tank using an Intel RealSense D455 camera. The dataset contains 1,068 grayscale images at a resolution of $640\times480$ for both the left and right stereo cameras. Stereo images and recorded IMU data are fed into OKVIS2~\cite{okvis2} to estimate camera poses for each left stereo image, though the estimated poses may contain errors. Leveraging recent advancements in depth estimation, we employ FoundationStereo~\cite{foundationstereo} to generate smoother and higher-quality depth maps (corresponding to the left stereo images) from the stereo pairs. The post-processed data, including left stereo images, depth maps, and camera poses, are then used as input to GSFusion~\cite{gsfusion} for the initial 3DGS reconstruction. Since no ground truth is available for this self-collected dataset, we randomly select five novel views in the reconstructed scene where shadow-like floaters caused by pose inaccuracies are prominent. For each, we use a nearby captured training view as a reference “ground truth” to qualitatively assess our method’s performance. Examples of the captured data and selected novel views are shown in \cref{fig:ship_cam0_vis} and \cref{fig:ship_cam0_vis_supp}. This real-world dataset will be released to support reproducible research.

\begin{table}[]
\centering
\resizebox{\columnwidth}{!}{%
\begin{tabular}{l|ccc}
\hlineB{2}
Method                         & PSNR$\uparrow$ & SSIM$\uparrow$ & LPIPS$\downarrow$ \\ \hlineB{2}
SplaTAM                        & 23.03          & 0.791          & 0.311             \\
SplaTAM + DIFIX                & 23.06          & 0.789          & 0.220             \\
SplaTAM + DIFIX-finetune$^\dagger$       & {\ul 24.67}    & {\ul 0.829}    & \textbf{0.133}    \\
SplaTAM + GSFixer              & \textbf{25.11} & \textbf{0.831} & {\ul 0.188}       \\ \hline
RTG-SLAM                       & 19.54          & 0.777          & 0.341             \\
RTG-SLAM + DIFIX               & 19.43          & 0.762          & 0.245             \\
RTG-SLAM + DIFIX-finetune$^\dagger$      & {\ul 24.44}    & {\ul 0.812}    & \textbf{0.156}    \\
RTG-SLAM + GSFixer             & \textbf{24.80} & \textbf{0.824} & {\ul 0.204}       \\ \hline
GSFusion (gs)                  & 24.58          & \textbf{0.838} & 0.308             \\
GSFusion (gs) + DIFIX          & 24.34          & 0.818          & 0.193             \\
GSFusion (gs) + DIFIX-finetune$^\dagger$ & {\ul 24.87}    & 0.834          & \textbf{0.142}    \\
GSFusion (gs) + GSFixer        & 24.79          & 0.833          & 0.196             \\
GSFusion (mesh+gs) + GSFixer   & \textbf{25.30} & {\ul 0.837}    & {\ul 0.183}       \\ \hlineB{2}
\end{tabular}%
}
\caption{More comparisons of diffusion-based repair methods on the ScanNet++ dataset. $^\dagger$ indicates that we fine-tuned this model based on the original DIFIX model. The best result is highlighted in \textbf{bold}, and the second-best is \uline{underlined}. The text inside ( ) indicates the format of the reconstruction used.}
\label{tab:supp_exp_scannetpp}
\end{table}

\begin{table}[]
\centering
\resizebox{\columnwidth}{!}{%
\begin{tabular}{l|ccc}
\hlineB{2}
Method                         & PSNR$\uparrow$ & SSIM$\uparrow$ & LPIPS$\downarrow$ \\ \hlineB{2}
SplaTAM                        & 23.82          & 0.833          & 0.267             \\
SplaTAM + DIFIX                & 22.97          & 0.790          & 0.262             \\
SplaTAM + DIFIX-finetune$^\dagger$       & {\ul 25.45}    & \textbf{0.867} & \textbf{0.149}    \\
SplaTAM + GSFixer              & \textbf{25.67} & {\ul 0.839}    & {\ul 0.215}       \\ \hline
RTG-SLAM                       & 25.00          & \textbf{0.860} & 0.247             \\
RTG-SLAM + DIFIX               & 24.02          & 0.811          & {\ul 0.214}       \\
RTG-SLAM + DIFIX-finetune$^\dagger$      & {\ul 25.47}    & {\ul 0.858}    & \textbf{0.156}    \\
RTG-SLAM + GSFixer             & \textbf{26.27} & 0.843          & 0.228             \\ \hline
GSFusion (gs)                  & 22.10          & 0.844          & 0.296             \\
GSFusion (gs) + DIFIX          & 21.81          & 0.772          & 0.273             \\
GSFusion (gs) + DIFIX-finetune$^\dagger$ & 23.12          & \textbf{0.847} & \textbf{0.185}    \\
GSFusion (gs) + GSFixer        & {\ul 23.87}    & 0.830          & 0.251             \\
GSFusion (mesh+gs) + GSFixer   & \textbf{25.98} & {\ul 0.845}    & {\ul 0.219}       \\ \hlineB{2}
\end{tabular}%
}
\caption{More comparisons of diffusion-based repair methods on the Replcia dataset. $^\dagger$ indicates that we fine-tuned this model based on the original DIFIX model. The best result is highlighted in \textbf{bold}, and the second-best is \uline{underlined}. The text inside ( ) indicates the format of the reconstruction used.}
\label{tab:supp_exp_replica}
\end{table}

\section{Additional Results}

\subsection{More DIFIX Variants}
\label{Sec:more_difix_results}
DIFIX and DIFIX-ref~\cite{difix} are diffusion models pretrained on 80k noisy-clean real image pairs created using their proposed dataset curation strategies, whereas our GSFixer is only pretrained on two synthetic datasets with randomly added Gaussian noise and blur, followed by fine-tuning on a small amount of clean captured data. To demonstrate the effectiveness and efficiency of our training strategy, we also fine-tune the original DIFIX model on the same scene data for 800 iterations on the ScanNet++ dataset and 400 iterations on the Replica dataset.

Due to the higher GPU memory demands of training DIFIX, we used an NVIDIA A40 GPU with 48GB VRAM to get DIFIX-finetune, while our GSFixer is trained on a 24GB NVIDIA RTX 4500 Ada GPU. Results are presented in \cref{tab:supp_exp_scannetpp} and \cref{tab:supp_exp_replica}. As expected, DIFIX-finetune shows improved performance over the original DIFIX on both datasets. However, it still falls short of GSFixer, particularly in PSNR, which correlates with lower visual quality and is clearly visible in \cref{fig:main_exp_vis_supp}. For instance, in the ScanNet++ dataset, the colorful floaters in the SplaTAM example are slightly suppressed by DIFIX-finetune, but not fully removed (also seen in the GSFusion example). In the RTG-SLAM example, DIFIX-finetune fills in the missing window area with content, but the result lacks the texture consistency achieved by GSFixer. Similarly, in the Replica dataset, although DIFIX-finetune reduces some artifacts compared to the original DIFIX, it still fails to inpaint large visible holes, which is an essential capability for novel view repair.

In conclusion, our fine-tuning protocol not only delivers superior performance in challenging scenarios but also requires significantly fewer computational resources and minimal dataset curation, making it both effective and efficient.

\begin{table}[]
\centering
\resizebox{\columnwidth}{!}{%
\begin{tabular}{l|l|ccc}
\hlineB{2}
Dataset                    & Method                       & PSNR$\uparrow$ & SSIM$\uparrow$ & LPIPS$\downarrow$ \\ \hlineB{2}
\multirow{6}{*}{ScanNet++} & SplaTAM                      & 23.03          & 0.791          & 0.311             \\
                           & SplaTAM (gs) + GSFixer       & {\ul 25.11}    & {\ul 0.831}    & {\ul 0.188}       \\
                           & SplaTAM (mesh$^*$+gs) + GSFixer  & \textbf{25.12} & \textbf{0.832} & \textbf{0.185}    \\ \cline{2-5} 
                           & RTG-SLAM                     & 19.54          & 0.777          & 0.341             \\
                           & RTG-SLAM (gs) + GSFixer      & {\ul 24.80}    & {\ul 0.824}    & {\ul 0.204}       \\
                           & RTG-SLAM (mesh$^*$+gs) + GSFixer & \textbf{25.05} & \textbf{0.827} & \textbf{0.191}    \\ \hline \hline
\multirow{6}{*}{Replica}   & SplaTAM                      & 23.82          & 0.833          & 0.267             \\
                           & SplaTAM (gs) + GSFixer       & {\ul 25.67}    & {\ul 0.839}    & {\ul 0.215}       \\
                           & SplaTAM (mesh$^*$+gs) + GSFixer  & \textbf{26.49} & \textbf{0.845} & \textbf{0.198}    \\ \cline{2-5} 
                           & RTG-SLAM                     & 25.00          & \textbf{0.860} & 0.247             \\
                           & RTG-SLAM (gs) + GSFixer      & {\ul 26.27}    & 0.843          & {\ul 0.228}       \\
                           & RTG-SLAM (mesh$^*$+gs) + GSFixer & \textbf{26.53} & {\ul 0.848}    & \textbf{0.212}    \\ \hlineB{2}
\end{tabular}%
}
\caption{More ablations of input image conditions on the ScanNet++ and Replica datasets. $^*$ denotes that the mesh reconstruction used for both SplaTAM and RTG-SLAM comparisons is borrowed from the GSFusion method.}
\label{tab:ablation_condition_supp}
\end{table}

\subsection{More Ablation Studies on Image Conditions}
\label{Sec:more_ablation}

\begin{figure*}[]
  \centering
   \includegraphics[width=\linewidth]{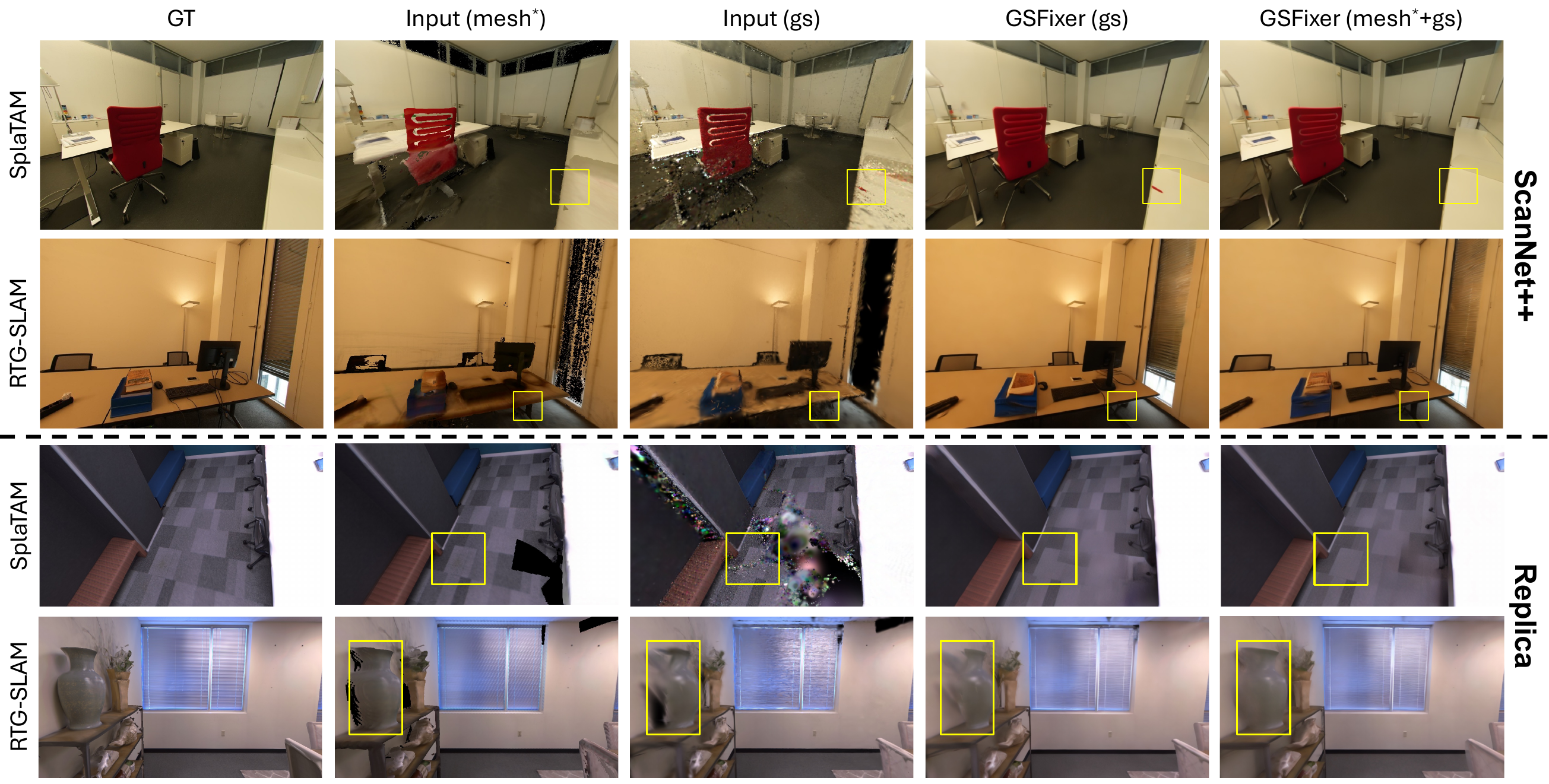}
   \caption{More qualitative ablations of input image conditions on the ScanNet++ and Replica datasets. The mesh reconstruction used for both SplaTAM and RTG-SLAM comparisons is borrowed from the GSFusion method. Zoom in to better observe how artifacts (highlighted by yellow boxes) present in the single-input settings are effectively mitigated with the dual-input configuration.}
   \label{fig:ablation_condition_vis_supp}
\end{figure*}

Although SplaTAM and RTG-SLAM do not produce meshes, we reuse the mesh extracted from GSFusion to render conditional images for fine-tuning and inference of GSFixer. As shown in \cref{tab:ablation_condition_supp}, GSFixer conditioned on dual-input consistently outperforms the single-input variant (conditioned only on 3DGS) across both datasets for SplaTAM and RTG-SLAM. These results reinforce the trends observed in \cref{Sec:ablation}. For instance, in the ScanNet++ dataset (see \cref{fig:ablation_condition_vis_supp}), floaters persist in the SplaTAM example and missing thin structures in the RTG-SLAM example are successfully corrected when mesh input is included. Similarly, in the Replica dataset (see \cref{fig:ablation_condition_vis_supp}), carpet textures in the SplaTAM example and the shape of the vase in the RTG-SLAM example are better preserved with the help of complementary information from the mesh input, which is otherwise lost in the 3DGS-only setting.

\begin{table}[]
\centering
\resizebox{\columnwidth}{!}{%
\begin{tabular}{l|l|ccc}
\hlineB{2}
Dataset                    & Method                           & PSNR$\uparrow$ & SSIM$\uparrow$ & LPIPS$\downarrow$ \\ \hlineB{2}
\multirow{6}{*}{ScanNet++} & SplaTAM                          & 23.03          & 0.791          & 0.311             \\
                           & SplaTAM (mesh$^*$+gs) + GSFixer  & {\ul 25.12}    & {\ul 0.832}    & \textbf{0.185}    \\
                           & SplaTAM (mesh$^*$+gs) + GSFix3D  & \textbf{25.21} & \textbf{0.836} & {\ul 0.218}       \\ \cline{2-5} 
                           & RTG-SLAM                         & 19.54          & 0.777          & 0.341             \\
                           & RTG-SLAM (mesh$^*$+gs) + GSFixer & {\ul 25.05}    & {\ul 0.827}    & \textbf{0.191}    \\
                           & RTG-SLAM (mesh$^*$+gs) + GSFix3D & \textbf{25.39} & \textbf{0.837} & {\ul 0.233}       \\ \hline \hline
\multirow{6}{*}{Replica}   & SplaTAM                          & 23.82          & 0.833          & 0.267             \\
                           & SplaTAM (mesh$^*$+gs) + GSFixer  & {\ul 26.49}    & {\ul 0.845}    & \textbf{0.198}    \\
                           & SplaTAM (mesh$^*$+gs) + GSFix3D  & \textbf{27.07} & \textbf{0.862} & {\ul 0.218}       \\ \cline{2-5} 
                           & RTG-SLAM                         & 25.00          & {\ul 0.860}    & 0.247             \\
                           & RTG-SLAM (mesh$^*$+gs) + GSFixer & {\ul 26.53}    & 0.848          & \textbf{0.212}    \\
                           & RTG-SLAM (mesh$^*$+gs) + GSFix3D & \textbf{27.18} & \textbf{0.868} & {\ul 0.236}       \\ \hlineB{2}
\end{tabular}%
}
\caption{More comparisons of GSFixer and GSFix3D on the ScanNet++ and Replica datasets. $^*$ denotes that the mesh reconstruction used for both SplaTAM and RTG-SLAM comparisons is borrowed from the GSFusion method.}
\label{tab:gsfix_exp_supp}
\end{table}

\subsection{More GSFix3D Comparisons}
\label{Sec:more_gsfix_results}

\begin{figure*}[]
  \centering
   \includegraphics[width=\linewidth]{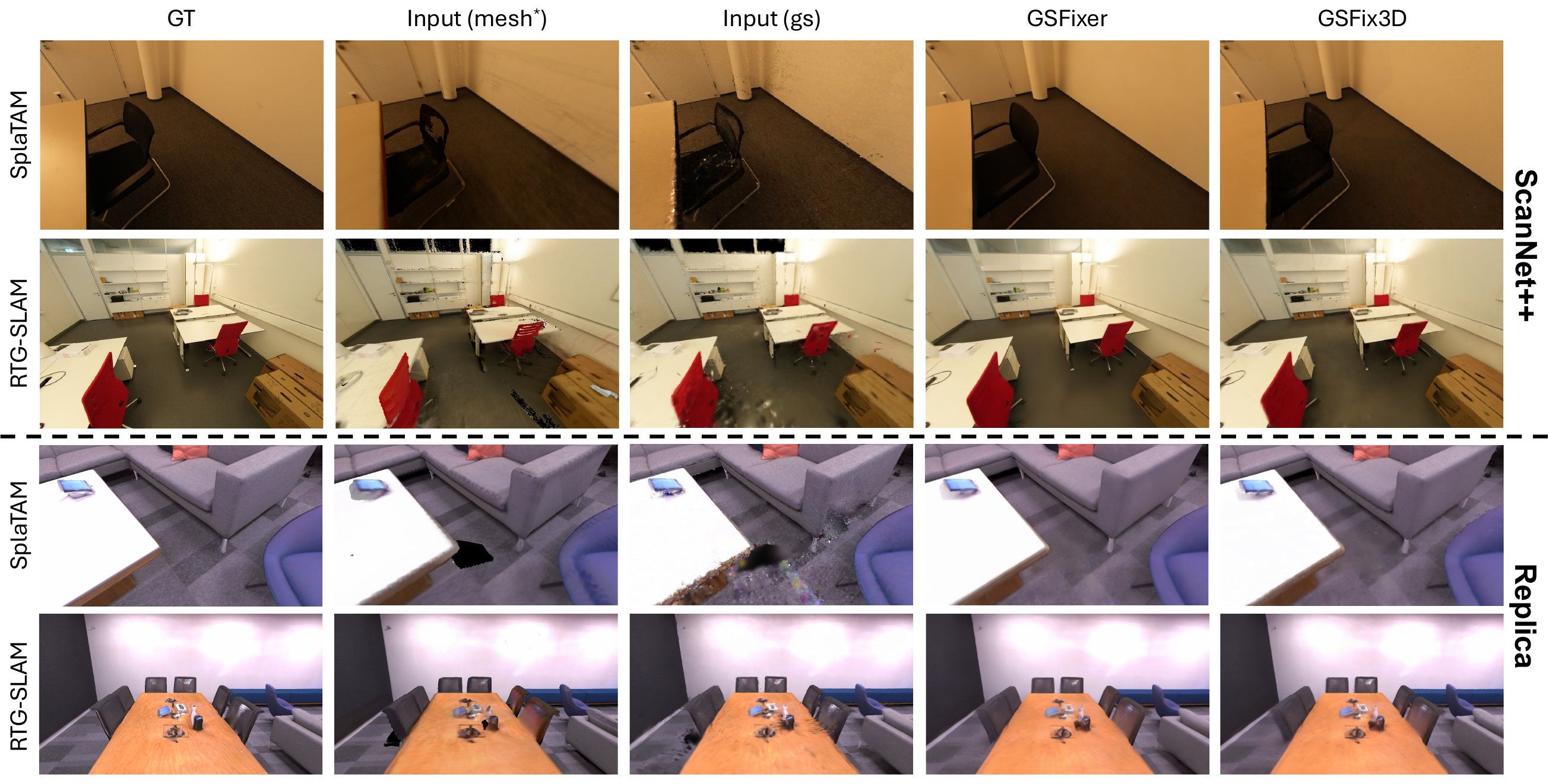}
   \caption{More qualitative comparisons between GSFixer and GSFix3D on the ScanNet++ and Replica dataset. Both mesh and 3DGS reconstructions are used as input sources. Zoom in to better observe how the 2D visual improvements from GSFixer are effectively distilled into the 3D space by GSFix3D.}
   \label{fig:gsfix_vis_supp}
\end{figure*}

To further demonstrate the flexibility of the GSFix3D framework, we incorporate reconstructions from SplaTAM and RTG-SLAM into our pipeline for novel view repair in 3D space. We reuse the extracted mesh from the GSFusion method to enable the dual-input setting, allowing us to fully leverage GSFixer's potential. The results in \cref{tab:gsfix_exp_supp} align with our main experiments on GSFusion presented in \cref{Sec:results}. The improvement in GSFix3D is attributed to the multi-view constraints applied during the optimization of 3D representations. Additional qualitative examples are provided in \cref{fig:gsfix_vis_supp}.

\subsection{More Real-World Tests}
\label{Sec:more_real_world}

\subsubsection{Self-collected Ship Data}
\label{Sec:more_ship_test}
\begin{figure*}[]
  \centering
   \includegraphics[width=0.7\linewidth]{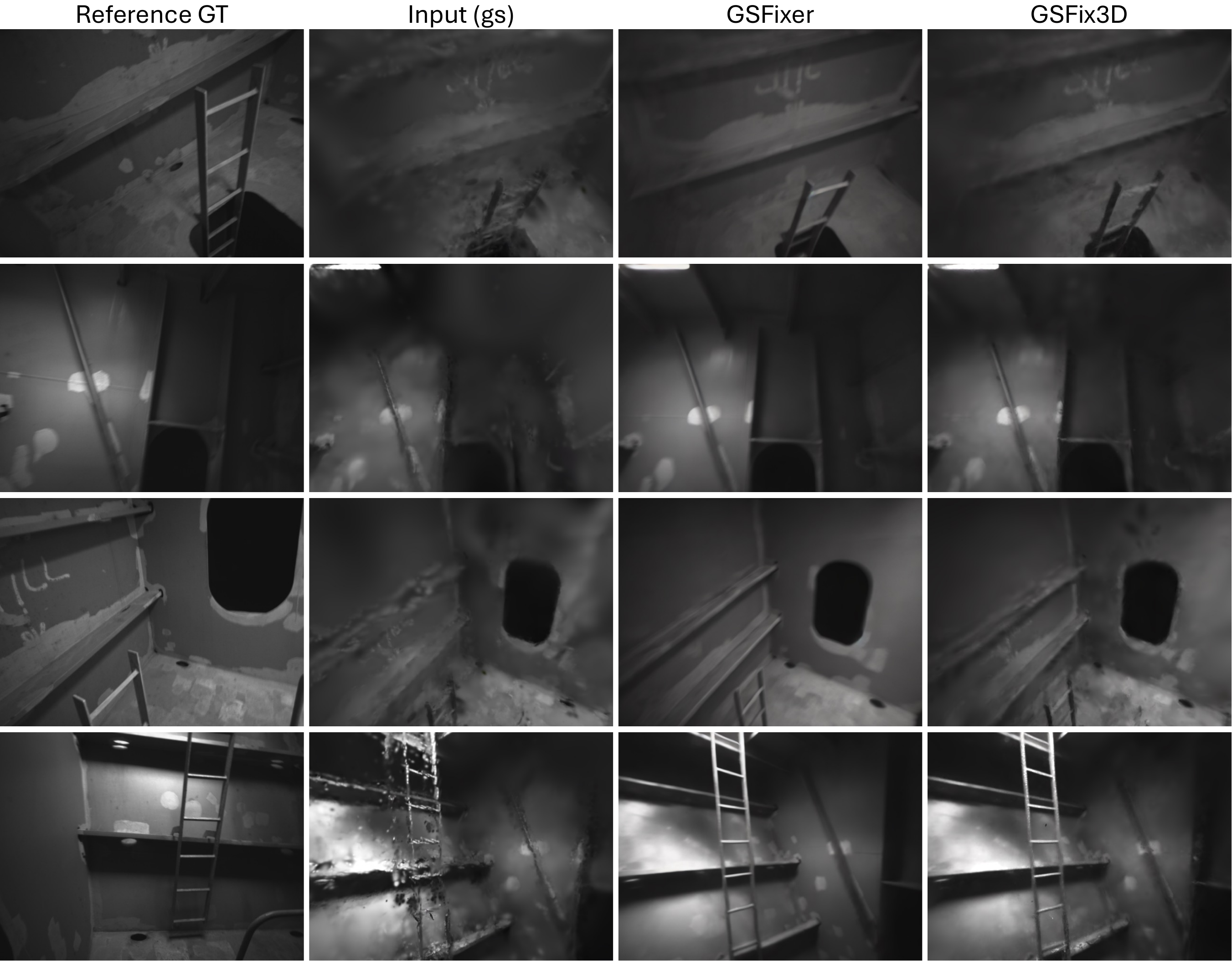}
   \caption{Novel view repair on self-collected ship data. Our method is robust to pose errors, effectively removing shadow-like floaters.}
   \label{fig:ship_cam0_vis_supp}
\end{figure*}

In \cref{fig:ship_cam0_vis_supp}, we present additional novel view repair results on our self-collected ship dataset. Since 3DGS is highly sensitive to pose inaccuracies, erroneous poses from multiple views can introduce shadow-like floaters in the scene, resulting in more severe artifacts in novel views. Despite this challenge, our methods, GSFixer and GSFix3D, successfully learn the artifact distribution from the captured training data through our proposed fine-tuning protocol, enabling effective removal in both the 2D image space and the 3D scene representation. This in-the-wild test further highlights the robustness of our approach.

\subsubsection{Outdoor Scenario}
\label{Sec:more_outdoor_test}
\begin{figure*}[]
  \centering
   \includegraphics[width=0.7\linewidth]{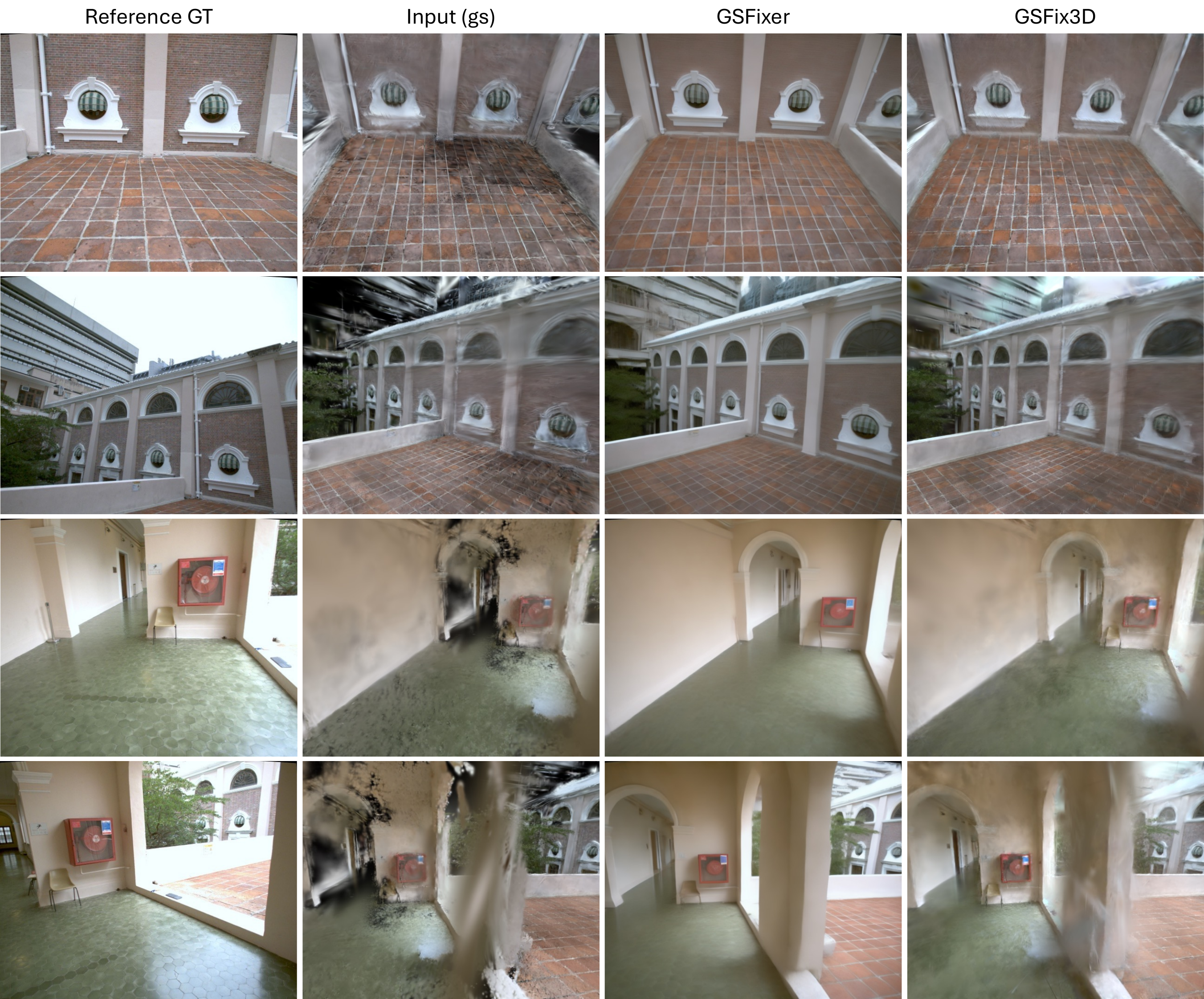}
   \caption{Novel view repair on a challenging outdoor scene from the FAST-LIVO dataset~\cite{fast-livo}. The initial reconstruction is generated by a LiDAR-Inertial-Camera Gaussian Splatting SLAM system~\cite{gaussian-lic}, which may contain pose errors and produce inaccurate maps. Our method manages to repair those broken geometries to some extent.}
   \label{fig:outdoor_lidar_vis}
\end{figure*}

To further demonstrate adaptability across scenes and reconstruction pipelines, we select a challenging real-world outdoor scene from the FAST-LIVO dataset~\cite{fast-livo}, covering building exteriors and archway corridors. The data is captured with a hard-synchronized LiDAR and camera setup, and we reconstruct the 3DGS scene using a LiDAR-Inertial-Camera Gaussian Splatting SLAM system~\cite{gaussian-lic} (Gaussian-LIC). Note that pose errors may still occur, producing inaccurate maps and broken geometries. We fine-tune GSFixer on the captured RGB data and the 3DGS renderings from Gaussian-LIC. As shown in \cref{fig:outdoor_lidar_vis}, our method manages to fill in visual holes on the brick ground, recover distant buildings and sky, and correct broken structures such as the arch wall and deep corridor, further validating GSFix3D’s robustness in challenging real-world scenarios.

%% file: main.bbl
\begin{thebibliography}{43}
\providecommand{\natexlab}[1]{#1}
\providecommand{\url}[1]{\texttt{#1}}
\expandafter\ifx\csname urlstyle\endcsname\relax
  \providecommand{\doi}[1]{doi: #1}\else
  \providecommand{\doi}{doi: \begingroup \urlstyle{rm}\Url}\fi

\bibitem[Cabon et~al.(2020)Cabon, Murray, and Humenberger]{vkitti}
Yohann Cabon, Naila Murray, and Martin Humenberger.
\newblock Virtual kitti 2.
\newblock \emph{arXiv preprint arXiv:2001.10773}, 2020.

\bibitem[Curless and Levoy(1996)]{tsdf}
Brian Curless and Marc Levoy.
\newblock A volumetric method for building complex models from range images.
\newblock In \emph{Proceedings of the 23rd annual conference on Computer graphics and interactive techniques}, pages 303--312, 1996.

\bibitem[Gu{\'e}don and Lepetit(2024)]{sugar}
Antoine Gu{\'e}don and Vincent Lepetit.
\newblock Sugar: Surface-aligned gaussian splatting for efficient 3d mesh reconstruction and high-quality mesh rendering.
\newblock In \emph{Proceedings of the IEEE/CVF Conference on Computer Vision and Pattern Recognition}, pages 5354--5363, 2024.

\bibitem[Ho et~al.(2020)Ho, Jain, and Abbeel]{ddpm}
Jonathan Ho, Ajay Jain, and Pieter Abbeel.
\newblock Denoising diffusion probabilistic models.
\newblock \emph{Advances in neural information processing systems}, 33:\penalty0 6840--6851, 2020.

\bibitem[Hu et~al.(2022)Hu, Shen, Wallis, Allen-Zhu, Li, Wang, Wang, Chen, et~al.]{lora}
Edward~J Hu, Yelong Shen, Phillip Wallis, Zeyuan Allen-Zhu, Yuanzhi Li, Shean Wang, Lu Wang, Weizhu Chen, et~al.
\newblock Lora: Low-rank adaptation of large language models.
\newblock \emph{ICLR}, 1\penalty0 (2):\penalty0 3, 2022.

\bibitem[Huang et~al.(2024)Huang, Yu, Chen, Geiger, and Gao]{2dgs}
Binbin Huang, Zehao Yu, Anpei Chen, Andreas Geiger, and Shenghua Gao.
\newblock 2d gaussian splatting for geometrically accurate radiance fields.
\newblock In \emph{ACM SIGGRAPH 2024 conference papers}, pages 1--11, 2024.

\bibitem[Ke et~al.(2025)Ke, Qu, Wang, Metzger, Huang, Li, Obukhov, and Schindler]{marigold}
Bingxin Ke, Kevin Qu, Tianfu Wang, Nando Metzger, Shengyu Huang, Bo Li, Anton Obukhov, and Konrad Schindler.
\newblock Marigold: Affordable adaptation of diffusion-based image generators for image analysis.
\newblock \emph{arXiv preprint arXiv:2505.09358}, 2025.

\bibitem[Keetha et~al.(2024)Keetha, Karhade, Jatavallabhula, Yang, Scherer, Ramanan, and Luiten]{splatam}
Nikhil Keetha, Jay Karhade, Krishna~Murthy Jatavallabhula, Gengshan Yang, Sebastian Scherer, Deva Ramanan, and Jonathon Luiten.
\newblock Splatam: Splat track \& map 3d gaussians for dense rgb-d slam.
\newblock In \emph{Proceedings of the IEEE/CVF Conference on Computer Vision and Pattern Recognition}, pages 21357--21366, 2024.

\bibitem[Kerbl et~al.(2023)Kerbl, Kopanas, Leimk{\"u}hler, and Drettakis]{3dgs}
Bernhard Kerbl, Georgios Kopanas, Thomas Leimk{\"u}hler, and George Drettakis.
\newblock 3d gaussian splatting for real-time radiance field rendering.
\newblock \emph{ACM Trans. Graph.}, 42\penalty0 (4):\penalty0 139--1, 2023.

\bibitem[Kingma et~al.(2013)Kingma, Welling, et~al.]{vae}
Diederik~P Kingma, Max Welling, et~al.
\newblock Auto-encoding variational bayes, 2013.

\bibitem[Lang et~al.(2024)Lang, Li, Wu, Zhao, Liu, Liu, Lv, and Zuo]{gaussian-lic}
Xiaolei Lang, Laijian Li, Chenming Wu, Chen Zhao, Lina Liu, Yong Liu, Jiajun Lv, and Xingxing Zuo.
\newblock Gaussian-lic: Real-time photo-realistic slam with gaussian splatting and lidar-inertial-camera fusion.
\newblock \emph{arXiv preprint arXiv:2404.06926}, 2024.

\bibitem[Leutenegger(2022)]{okvis2}
Stefan Leutenegger.
\newblock Okvis2: Realtime scalable visual-inertial slam with loop closure.
\newblock \emph{arXiv preprint arXiv:2202.09199}, 2022.

\bibitem[Liu et~al.(2024{\natexlab{a}})Liu, Chen, Kao, Tai, and Tang]{deceptivenerf}
Xinhang Liu, Jiaben Chen, Shiu-Hong Kao, Yu-Wing Tai, and Chi-Keung Tang.
\newblock Deceptive-nerf/3dgs: Diffusion-generated pseudo-observations for high-quality sparse-view reconstruction.
\newblock In \emph{European Conference on Computer Vision}, pages 337--355. Springer, 2024{\natexlab{a}}.

\bibitem[Liu et~al.(2024{\natexlab{b}})Liu, Zhou, and Huang]{3dgsenhancer}
Xi Liu, Chaoyi Zhou, and Siyu Huang.
\newblock 3dgs-enhancer: Enhancing unbounded 3d gaussian splatting with view-consistent 2d diffusion priors.
\newblock \emph{Advances in Neural Information Processing Systems}, 37:\penalty0 133305--133327, 2024{\natexlab{b}}.

\bibitem[Matsuki et~al.(2024)Matsuki, Murai, Kelly, and Davison]{monogs}
Hidenobu Matsuki, Riku Murai, Paul~HJ Kelly, and Andrew~J Davison.
\newblock Gaussian splatting slam.
\newblock In \emph{Proceedings of the IEEE/CVF Conference on Computer Vision and Pattern Recognition}, pages 18039--18048, 2024.

\bibitem[Mildenhall et~al.(2021)Mildenhall, Srinivasan, Tancik, Barron, Ramamoorthi, and Ng]{nerf}
Ben Mildenhall, Pratul~P Srinivasan, Matthew Tancik, Jonathan~T Barron, Ravi Ramamoorthi, and Ren Ng.
\newblock Nerf: Representing scenes as neural radiance fields for view synthesis.
\newblock \emph{Communications of the ACM}, 65\penalty0 (1):\penalty0 99--106, 2021.

\bibitem[Mou et~al.(2024)Mou, Wang, Xie, Wu, Zhang, Qi, and Shan]{t2i}
Chong Mou, Xintao Wang, Liangbin Xie, Yanze Wu, Jian Zhang, Zhongang Qi, and Ying Shan.
\newblock T2i-adapter: Learning adapters to dig out more controllable ability for text-to-image diffusion models.
\newblock In \emph{Proceedings of the AAAI conference on artificial intelligence}, pages 4296--4304, 2024.

\bibitem[Newcombe et~al.(2011)Newcombe, Izadi, Hilliges, Molyneaux, Kim, Davison, Kohi, Shotton, Hodges, and Fitzgibbon]{kinectfusion}
Richard~A Newcombe, Shahram Izadi, Otmar Hilliges, David Molyneaux, David Kim, Andrew~J Davison, Pushmeet Kohi, Jamie Shotton, Steve Hodges, and Andrew Fitzgibbon.
\newblock Kinectfusion: Real-time dense surface mapping and tracking.
\newblock In \emph{2011 10th IEEE international symposium on mixed and augmented reality}, pages 127--136. Ieee, 2011.

\bibitem[Nichol et~al.(2021)Nichol, Dhariwal, Ramesh, Shyam, Mishkin, McGrew, Sutskever, and Chen]{dm2021}
Alex Nichol, Prafulla Dhariwal, Aditya Ramesh, Pranav Shyam, Pamela Mishkin, Bob McGrew, Ilya Sutskever, and Mark Chen.
\newblock Glide: Towards photorealistic image generation and editing with text-guided diffusion models.
\newblock \emph{arXiv preprint arXiv:2112.10741}, 2021.

\bibitem[Nie{\ss}ner et~al.(2013)Nie{\ss}ner, Zollh{\"o}fer, Izadi, and Stamminger]{voxelhasing}
Matthias Nie{\ss}ner, Michael Zollh{\"o}fer, Shahram Izadi, and Marc Stamminger.
\newblock Real-time 3d reconstruction at scale using voxel hashing.
\newblock \emph{ACM Transactions on Graphics (ToG)}, 32\penalty0 (6):\penalty0 1--11, 2013.

\bibitem[Oleynikova et~al.(2017)Oleynikova, Taylor, Fehr, Siegwart, and Nieto]{voxblox}
Helen Oleynikova, Zachary Taylor, Marius Fehr, Roland Siegwart, and Juan Nieto.
\newblock Voxblox: Incremental 3d euclidean signed distance fields for on-board mav planning.
\newblock In \emph{2017 IEEE/RSJ International Conference on Intelligent Robots and Systems (IROS)}, pages 1366--1373. IEEE, 2017.

\bibitem[Paliwal et~al.(2025)Paliwal, Zhou, Ye, Xiong, Ranjan, and Kalantari]{ri3d}
Avinash Paliwal, Xilong Zhou, Wei Ye, Jinhui Xiong, Rakesh Ranjan, and Nima~Khademi Kalantari.
\newblock Ri3d: Few-shot gaussian splatting with repair and inpainting diffusion priors.
\newblock 2025.

\bibitem[Peng et~al.(2024)Peng, Shao, Liu, Zhou, Yang, Wang, and Zhou]{rtgslam}
Zhexi Peng, Tianjia Shao, Yong Liu, Jingke Zhou, Yin Yang, Jingdong Wang, and Kun Zhou.
\newblock Rtg-slam: Real-time 3d reconstruction at scale using gaussian splatting.
\newblock In \emph{ACM SIGGRAPH 2024 Conference Papers}, pages 1--11, 2024.

\bibitem[Roberts et~al.(2021)Roberts, Ramapuram, Ranjan, Kumar, Bautista, Paczan, Webb, and Susskind]{hypersim}
Mike Roberts, Jason Ramapuram, Anurag Ranjan, Atulit Kumar, Miguel~Angel Bautista, Nathan Paczan, Russ Webb, and Joshua~M Susskind.
\newblock Hypersim: A photorealistic synthetic dataset for holistic indoor scene understanding.
\newblock In \emph{Proceedings of the IEEE/CVF international conference on computer vision}, pages 10912--10922, 2021.

\bibitem[Rombach et~al.(2022)Rombach, Blattmann, Lorenz, Esser, and Ommer]{stablediffusion}
Robin Rombach, Andreas Blattmann, Dominik Lorenz, Patrick Esser, and Bj{\"o}rn Ommer.
\newblock High-resolution image synthesis with latent diffusion models.
\newblock In \emph{Proceedings of the IEEE/CVF conference on computer vision and pattern recognition}, pages 10684--10695, 2022.

\bibitem[Song et~al.(2020)Song, Meng, and Ermon]{ddim}
Jiaming Song, Chenlin Meng, and Stefano Ermon.
\newblock Denoising diffusion implicit models.
\newblock \emph{arXiv preprint arXiv:2010.02502}, 2020.

\bibitem[Song and Ermon(2019)]{dm2019}
Yang Song and Stefano Ermon.
\newblock Generative modeling by estimating gradients of the data distribution.
\newblock \emph{Advances in neural information processing systems}, 32, 2019.

\bibitem[Steinbrucker et~al.(2013)Steinbrucker, Kerl, and Cremers]{steinbrucker2013large}
Frank Steinbrucker, Christian Kerl, and Daniel Cremers.
\newblock Large-scale multi-resolution surface reconstruction from rgb-d sequences.
\newblock In \emph{Proceedings of the IEEE International Conference on Computer Vision}, pages 3264--3271, 2013.

\bibitem[Straub et~al.(2019)Straub, Whelan, Ma, Chen, Wijmans, Green, Engel, Mur-Artal, Ren, Verma, et~al.]{replica}
Julian Straub, Thomas Whelan, Lingni Ma, Yufan Chen, Erik Wijmans, Simon Green, Jakob~J Engel, Raul Mur-Artal, Carl Ren, Shobhit Verma, et~al.
\newblock The replica dataset: A digital replica of indoor spaces.
\newblock \emph{arXiv preprint arXiv:1906.05797}, 2019.

\bibitem[Sucar et~al.(2021)Sucar, Liu, Ortiz, and Davison]{imap}
Edgar Sucar, Shikun Liu, Joseph Ortiz, and Andrew~J Davison.
\newblock imap: Implicit mapping and positioning in real-time.
\newblock In \emph{Proceedings of the IEEE/CVF international conference on computer vision}, pages 6229--6238, 2021.

\bibitem[Vespa et~al.(2018)Vespa, Nikolov, Grimm, Nardi, Kelly, and Leutenegger]{supereight}
Emanuele Vespa, Nikolay Nikolov, Marius Grimm, Luigi Nardi, Paul~HJ Kelly, and Stefan Leutenegger.
\newblock Efficient octree-based volumetric slam supporting signed-distance and occupancy mapping.
\newblock \emph{IEEE Robotics and Automation Letters}, 3\penalty0 (2):\penalty0 1144--1151, 2018.

\bibitem[Wasserman et~al.(2025)Wasserman, Rotstein, Ganz, and Kimmel]{pipe_masks}
Navve Wasserman, Noam Rotstein, Roy Ganz, and Ron Kimmel.
\newblock Paint by inpaint: Learning to add image objects by removing them first.
\newblock In \emph{Proceedings of the Computer Vision and Pattern Recognition Conference}, pages 18313--18324, 2025.

\bibitem[Wei and Leutenegger(2024)]{gsfusion}
Jiaxin Wei and Stefan Leutenegger.
\newblock Gsfusion: Online rgb-d mapping where gaussian splatting meets tsdf fusion.
\newblock \emph{IEEE Robotics and Automation Letters}, 2024.

\bibitem[Wen et~al.(2025)Wen, Trepte, Aribido, Kautz, Gallo, and Birchfield]{foundationstereo}
Bowen Wen, Matthew Trepte, Joseph Aribido, Jan Kautz, Orazio Gallo, and Stan Birchfield.
\newblock Foundationstereo: Zero-shot stereo matching.
\newblock In \emph{Proceedings of the Computer Vision and Pattern Recognition Conference}, pages 5249--5260, 2025.

\bibitem[Wu et~al.(2025{\natexlab{a}})Wu, Zhang, Turki, Ren, Gao, Shou, Fidler, Gojcic, and Ling]{difix}
Jay~Zhangjie Wu, Yuxuan Zhang, Haithem Turki, Xuanchi Ren, Jun Gao, Mike~Zheng Shou, Sanja Fidler, Zan Gojcic, and Huan Ling.
\newblock Difix3d+: Improving 3d reconstructions with single-step diffusion models.
\newblock In \emph{Proceedings of the Computer Vision and Pattern Recognition Conference}, pages 26024--26035, 2025{\natexlab{a}}.

\bibitem[Wu et~al.(2025{\natexlab{b}})Wu, Xu, Huang, Geiger, and Chen]{genfusion}
Sibo Wu, Congrong Xu, Binbin Huang, Andreas Geiger, and Anpei Chen.
\newblock Genfusion: Closing the loop between reconstruction and generation via videos.
\newblock In \emph{Proceedings of the Computer Vision and Pattern Recognition Conference}, pages 6078--6088, 2025{\natexlab{b}}.

\bibitem[Yan et~al.(2024)Yan, Qu, Xu, Zhao, Wang, Wang, and Li]{gsslam}
Chi Yan, Delin Qu, Dan Xu, Bin Zhao, Zhigang Wang, Dong Wang, and Xuelong Li.
\newblock Gs-slam: Dense visual slam with 3d gaussian splatting.
\newblock In \emph{Proceedings of the IEEE/CVF Conference on Computer Vision and Pattern Recognition}, pages 19595--19604, 2024.

\bibitem[Yeshwanth et~al.(2023)Yeshwanth, Liu, Nie{\ss}ner, and Dai]{scannetpp}
Chandan Yeshwanth, Yueh-Cheng Liu, Matthias Nie{\ss}ner, and Angela Dai.
\newblock Scannet++: A high-fidelity dataset of 3d indoor scenes.
\newblock In \emph{Proceedings of the IEEE/CVF International Conference on Computer Vision}, pages 12--22, 2023.

\bibitem[Yu et~al.(2025)Yu, Wang, Yang, Wang, Cao, Ji, and Sun]{sgd}
Zhongrui Yu, Haoran Wang, Jinze Yang, Hanzhang Wang, Jiale Cao, Zhong Ji, and Mingming Sun.
\newblock Sgd: Street view synthesis with gaussian splatting and diffusion prior.
\newblock In \emph{2025 IEEE/CVF Winter Conference on Applications of Computer Vision (WACV)}, pages 3812--3822. IEEE, 2025.

\bibitem[Zhang et~al.(2023)Zhang, Rao, and Agrawala]{controlnet}
Lvmin Zhang, Anyi Rao, and Maneesh Agrawala.
\newblock Adding conditional control to text-to-image diffusion models.
\newblock In \emph{Proceedings of the IEEE/CVF international conference on computer vision}, pages 3836--3847, 2023.

\bibitem[Zheng et~al.(2022)Zheng, Zhu, Xu, Liu, Guo, and Zhang]{fast-livo}
Chunran Zheng, Qingyan Zhu, Wei Xu, Xiyuan Liu, Qizhi Guo, and Fu Zhang.
\newblock Fast-livo: Fast and tightly-coupled sparse-direct lidar-inertial-visual odometry.
\newblock In \emph{2022 IEEE/RSJ international conference on intelligent robots and systems (IROS)}, pages 4003--4009. IEEE, 2022.

\bibitem[Zhong et~al.(2025)Zhong, Li, Chen, Hong, and Xu]{tamingvideo}
Yingji Zhong, Zhihao Li, Dave~Zhenyu Chen, Lanqing Hong, and Dan Xu.
\newblock Taming video diffusion prior with scene-grounding guidance for 3d gaussian splatting from sparse inputs.
\newblock In \emph{Proceedings of the Computer Vision and Pattern Recognition Conference}, pages 6133--6143, 2025.

\bibitem[Zhu et~al.(2022)Zhu, Peng, Larsson, Xu, Bao, Cui, Oswald, and Pollefeys]{niceslam}
Zihan Zhu, Songyou Peng, Viktor Larsson, Weiwei Xu, Hujun Bao, Zhaopeng Cui, Martin~R Oswald, and Marc Pollefeys.
\newblock Nice-slam: Neural implicit scalable encoding for slam.
\newblock In \emph{Proceedings of the IEEE/CVF conference on computer vision and pattern recognition}, pages 12786--12796, 2022.

\end{thebibliography}
